\documentclass{ieeeaccess}
\usepackage{cite}
\usepackage{amsmath,amssymb,amsfonts}
\usepackage{algorithmic}
\usepackage{graphicx}
\usepackage{textcomp}

\usepackage{cuted} 
\usepackage[font={sf,scriptsize},           
labelfont={bf,color=accessblue},
caption=false]                  
{subfig} 

\usepackage{amsthm}       
{
	\theoremstyle{plain}
	\newtheorem{assumption}{Assumption}
	\newtheorem{corollary}{Corollary}
}
\def\BibTeX{{\rm B\kern-.05em{\sc i\kern-.025em b}\kern-.08em
    T\kern-.1667em\lower.7ex\hbox{E}\kern-.125emX}}
\begin{document}
\history{© 2022 IEEE. Personal use of this material is permitted. 
Permission from IEEE must be obtained for all other uses, including reprinting/republishing this material	for advertising or promotional purposes, collecting new collected works for resale 	or redistribution to servers or lists, or reuse of any copyrighted component of this work in other works. \newline 
This work has been submitted to the IEEE for possible publication. Copyright may be transferred without notice, after which this version
may no longer be accessible.
Please cite the newer, accepted version that has the DOI below:}
\doi{10.1109/ACCESS.2022.3169766}

\title{Analytic Solutions for Wheeled Mobile Manipulator Supporting Forces}
\author{\uppercase{Goran R. Petrovi{\'c}}\authorrefmark{1}
\uppercase{and Jouni Mattila}\authorrefmark{1}}
\address[1]{Faculty of Engineering and Natural Sciences, Unit of Automation Technology and Mechanical Engineering, Tampere University, Tampere, Finland}
\tfootnote{This project STREAM has received funding from the Shift2Rail joint Undertaking (JU) under grant agreement No. 101015418. The JU receives support from the European Union's Horizon 2020 research and innovation programme and the Shift2Rail JU members other than the Union. The content of this paper does not reflect the official opinion of the Shift2Rail Joint Undertaking (S2R JU). Responsibility for the information and views expressed in the paper lies entirely with the authors.}

\markboth
{Goran R. Petrovi{\'c} \headeretal: Analytic solutions for wheeled mobile manipulator supporting forces}
{Goran R. Petrovi{\'c} \headeretal: Analytic solutions for wheeled mobile manipulator supporting forces}

\corresp{Corresponding author: Goran R. Petrovi{\'c} (e-mail: goran.petrovic@tuni.fi).}

\begin{abstract}
When a mobile manipulator's wheel loses contact with the ground, tipping-over may occur, causing material damage, and in the worst case, it can put human lives in danger. The tip-over stability of wheeled mobile manipulators must not be overlooked at any stage of a mobile manipulator's life, starting from the design phase, continuing through the commissioning period and extending to the operational phase. Many tip-over stability criteria formulated throughout the years do not explicitly consider the normal wheel loads, with most of them relying on prescribed stability margins in terms of overturning moments. In these formulations, it is commonly argued that overturning will occur about one of the axes connecting adjacent manipulator's contact points with the ground. This claim may not always be valid and is certainly restrictive. Explicit expressions for the manipulator supporting forces provide the best insight into relevant affecting terms which contribute to the tip-over (in)stability. They also remove the necessity for thinking about which axis the manipulator could tip over and simultaneously enable the formulation of more intuitive stability margins and on-line tip-over prevention techniques. The present study presents a general dynamics modelling approach in the Newton--Euler framework using 6D vectors and gives normal wheel load equations in a typical 4-wheeled mobile manipulator negotiating a slope. The given expressions are expected to become standard in wheeled mobile manipulators and to provide a basis for effective tip-over stability criteria and tip-over avoidance techniques. Based on the presented results, specific improvements of the state-of-the-art criteria are discussed.
\end{abstract}


\titlepgskip=-40pt

\maketitle

\section{Introduction}

\label{sec:introduction}

\PARstart{M}{obile} manipulators have considerable application potential in various fields such as mining, logging, construction, earth-moving, searching and rescuing, agriculture, and planetary exploration, among others, \cite{b1}--\cite{b2}. Tipping-over stability of these vehicles equipped with a manipulator arm is crucial regardless of the level of automation. Heavy-duty mobile manipulators are in danger of tipping over if operated inadequately by an unskilled (tele)operator and in case of unexpected occurrences. Slope negotiations in sites where soil may also be unstable can be incredibly challenging \cite{b3}. In dexterous, manoeuvrable, compact and lightweight robots, the tipping-over danger is due to a high centre of mass (COM), small weight and  ground base, especially in combination with high loads and accelerations \cite{b4}.

Simple thought experiments can suffice to introduce the motivation for the following analysis. Let us consider, for example, a lightweight wheeled platform that carries a manipulator arm of comparable weight. It is intuitively clear that overturning can occur in specific disadvantageous arm postures, especially with external forces acting on the manipulator tip, as when carrying a load or interacting with surroundings. An additional aggravating factor, in this case, can be sloped and rugged terrain. The tipping-over starts by making one or more ground reactions equal to zero at first, and then if some preventive action is not taken on time, material and collateral damage are imminent. Extra special care must be taken if there exists a manipulator-human interaction. In other \textit{mise-en-sc{\`e}ne}, where a human-operated heavy-duty mobile manipulator (e.g., an excavator) is working in an environment with other people nearby, human lives can be at stake in a severe turn of events if tipping-over occurs. These short considerations in two different settings also emphasise the significance of monitoring tyre loads even in semi- and non-automated solutions, making tip-over stability indicators necessary in regular everyday use. However, in the absence of (tele)operator action, fully-automated solutions rely wholly on how good the tip-over stability indicator is formulated. 

The tip-over analysis of a mobile manipulator should come into focus, starting from the early design stages where the number of wheels, size, mass, the position of the manipulator arm and possibly other manipulator parameters are optimised to maximise the workspace and provide the relative stability margin against tipping over in the most critical predicted cases \cite{b5}. Even detailed analyses like this, performed in advance, may not anticipate the specific, unexpected course of events during the operation. Thus, tipping-over monitoring should create an alert and, favourably, start the tip-over avoidance sequence whenever the tip-over danger exists.

A well-defined tipping-over stability indicator may explicitly or implicitly address the wheel supporting forces. It must also include all the relevant factors that affect them. Although the relevance of each one, \textit{per se}, might not be the same from case to case, all the potentially influential factors must be preliminarily considered in a general contemplation. These are all the system masses and moments of inertia, together with all the significant linear/angular accelerations/velocities/positions and terrain slope.

As the literature review in Section \ref{sec: literature} shows, tip-over stability monitoring in wheeled mobile manipulators has received significant attention continuously over the years. Despite various proposed approaches with different underlying concepts and modelling complexity, the methods lack detailed expressions for normal loads and tip-over stability criteria considering supporting forces accompanied by full-dynamics modelling. It must be duly noted that for some reason, a gap existed between the car and mobile manipulators field of research, although weight transfer to tyres is a common sphere of interest. Once this gap is closed and expanded expressions for normal tyre loads are derived from the full-dynamics model, all the assumptions on overturning axis location will not be significant. The tipping-over criterion can be formulated most naturally in terms of normal loads. By monitoring the normal wheel loads, one can effectively trace if the value of any supporting forces approaches some prescribed critical value, which is an intuitive problem solution.

\textit{As the main contribution}, the presented study closes the existing gap between the car and mobile manipulator dynamics by providing an extension of expressions for normal car wheel loads to the case where a manipulator arm exists on top of a wheeled platform. It is further shown how the wheel-loads-based tip-over stability indicator outperforms the widespread moments-based indicators. Appropriate alternatives to the state-of-the-art criteria are suggested.

The present study provides a detailed dynamics model, under the veil of the {N--E} formalism using 6D vectors, of a 4-AWD (all-wheel drive) mobile manipulator negotiating a slope. The specimen 4-wheeled AWD mobile manipulator is chosen for analysis since it offers a fair amount of generality without introducing excessive complexity. Heavy-duty machines with Ackermann or skid steering can be seen as special cases of the case presented here. Additional efforts must be made in the case of articulated steering and case with six or more wheels. Since the same reasoning and line of thought as presented here should be followed in those situations, they will not receive special attention in the following analysis. Subsystem-by-subsystem modelling and underlying analysis have been carried out in detail with a minimum number of reasonable assumptions to balance the complexity and practicality  with the modelling accuracy.

Expressions for normal wheel loads are derived in a  neat and structured manner. These provide the basis for the proposed Tipping Over Force (TOF) criterion, which does not rely on a typically used tipping-over axis, and can be seen as a quick and better alternative to the state-of-the-art criteria. It is expected that this or similar criteria will become a de-facto standard tip-over stability indicator in mobile manipulators. Having explicit expressions for normal wheel loads makes it easy to carry on a term by term analysis starting from the design phase. It also makes the formulation of tipping-over prevention actions more straightforward. Apart from getting a good general insight, end-users will also be able to tailor the given expression according to their own needs.  

The validity of derived expressions for normal forces is advocated in the  Simscape Multibody\texttrademark \, by comparing results to the unbiased reference from the renowned software. 

The rest of the paper is organised as follows. Section \ref{sec: literature} gives a literature overview to situate the present study better. Section \ref{sec:formalism} motivates the use of the {N--E} formalism in the discussed context. Section \ref{sec: preliminaries} presents essential mathematical preliminaries. Section \ref{sec: kinematic chain} addresses the kinematics of a mobile manipulator negotiating a  slope. Section  \ref{sec: 6d dyn} provides 6D vector models of the wheel and chassis dynamics. Section \ref{sec: solution} deals with equations whose solutions are the tyre supporting forces. Section \ref{sec: simulation} presents the simulation results and suggests improvements of relevant tip-over stability and avoidance criteria. Section \ref{sec: discussion} contemplates the obtained results. Section \ref{sec: conclusions} summarises the conclusions drawn.

\section{Literature review}

\label{sec: literature}

The idea of Zero Moment Point (ZMP), \cite{b6}, addresses a point on the ground where the resultant moment of the external and inertial forces is equal to zero. Initially proposed in \cite{b7} for use in mobile manipulators, referring to the position of the ZMP with respect to the stability polygon, it has remained present in the research of tipping-over stability and related topics. Often, remarkably simplified dynamics models are combined with ZMP, and this fact has been a common cause for criticism in the mobile-manipulators community. Using the ZMP-based stability criterion from paper \cite{b8}, paper \cite{b9} reports the ZMP as a less sensitive indicator than Force-Angle (FA) or Moment-Height Stability (MHS) in certain mobile manipulators.  However, ZMP has proven to be helpful in on-line trajectory planning both for light and heavy mobile manipulators \cite{b10} -- \cite{b11} and for quadrupedal ones with wheels \cite{b12}. It contrasts the full-dynamics modelling narrative promoted here, with the primary aim being deriving the expressions for supporting forces.

Continuing efforts on developing improved tip-over indicators gave rise to the FA indicator \cite{b13}, with heavy-duty mobile machines serving as the primary source of inspiration. It is noticed that this criterion would provide a relevant and reliable indication at low speeds and with external forces of large magnitudes. The FA stability indicator considers the angle between the net force (excluding the ground support reaction forces) reduced to the planar mass system's COM and the rays connecting the same COM with the ground connection points. The vital part of the discussion concerns the tipping-over axis. \textit{Natural tip-over} and \textit{tripped tip-over} notions have been introduced. The tip-over was named {natural} if the overturning occurs about one of the axes connecting manipulator support points in contact with the ground (support polygon vertices). The {tripped} tip-over occurs about an axis representing a linear combination of the abovementioned axes. Examining the tripped tip-over seems to be abandoned in the research mainstream, and it will be recalled here. FA tip-over prevention algorithm was presented in \cite{b14}.

In \cite{b15}, also assuming that tipping-over will occur about one of the axes connecting the manipulator wheels, the MHS stability criterion was formulated utilising overturning moments without explicitly addressing ground reactions. A simple means to include the chassis COM height were presented and created a significant impact. Being a direct consequence of the dynamics modelling in the Newton--Euler (N--E) framework, it has neatly brought focus to forces/moments acting at chassis/manipulator base connection. The formulation has opened a path to more detailed stability criteria by explicitly addressing certain key factors in the normal load analysis. An on-line tip-over prevention MHS-based criterion was proposed in \cite{b16} and compared to the FA-method based.

Paper \cite{b17} resumes the established trend of formulating tipping-over stability indicators using moments about the manipulator supporting polygon axes. The Tipping Over Moment (TOM) criterion is an extended and improved version of the MHS. It includes the wheeled platform weight in the analysis and presents a reasonably formulated criterion, which, similarly to its predecessor MHS, does not explicitly consider the wheeled platform inertial forces and ground reactions. The TOM essentially investigates the values of the anti-tipping-over moment, referring to the negative moment values as the ones providing stability. These moments are compared to the prescribed relative stability margin values, as is done in paper \cite{b18} for a dual-armed wheel robot.

By examining the research trends in the tip-over stability of mobile manipulators, an impression is that there is a striving to improve TOM by performing modelling with fewer approximations in a usable manner. Paper \cite{b19} introduces a significant Improved Tipping Over Moment (ITOM) indicator. This indicator is indeed qualified to be named like that since it brings a wheeled platform's inertial forces to TOM. They have been one of the usually neglected terms throughout the years in this line of research, although they are undoubtedly worth considering, having in mind a significant dynamic coupling between the manipulator arm and the wheeled platform. The analysis which led to the ITOM formulation considered a manipulator negotiating a constant slope in the direction of motion. As with TOM, tipping-over axes with ITOM were again assumed to connect vertices of the support polygon and thus, supporting forces were not considered explicitly. It was also argued that analytic expressions for ITOM are hard to obtain. A complex manipulator arm inevitably leads to complicated equations of motion, but with careful rearrangement in a structured manner, many significant insights can be obtained from neat analytic expressions. Although specific challenges exist, it is plausible to obtain relatively simple analytic expressions for supporting forces by expanding ideas from car dynamics.

A detailed discussion regarding terrain slope has usually been avoided when the tip-over stability was examined. Commonly, two Euler angles at most were sufficient to describe the chassis orientation  to the inertial frame of reference, fixed in the Earth-tangent plane. Interestingly enough, weight transfer to wheels of an accelerating platform negotiating a slope has been an omnipresent topic in car dynamics, \cite{b20}. 

Weight transfer to the wheels, i.e. the tyre normal load, had always required special attention in the field of car dynamics since the standard tyre/road interaction models use the normal force in expressions for tyre/road forces/moments, \cite{b21}-- \cite{b22}. Thus, this issue has been recognised and widely addressed, usually providing approximate expressions which are sufficient to address the car motion. The problem of quantifying wheel loads becomes complicated in mobile manipulators with a manipulator arm  attached to the chassis and interacting with surroundings. In the general case, where the tyre weight can not be neglected, and both manipulator posture and movements will affect the normal tyre load, existing straightforward expressions require an extension. 

An effort in explicitly formulating tipping-over stability criterion using wheel loads can be found in \cite{b23}, where the analysis of the dynamics is over-simplified. Explicit formulation using normal wheel loads is experimentally addressed in \cite{b24} on a small-scale laboratory test bench. However, since the discussed approach requires measuring normal loads, it falls out of the perspective for the proposed narrative because of high cost and potentially impossible force sensor integrity preservation in heavy-duty mobile manipulators. Explicit formulation of the stability criterion in terms of wheel loads for a 3-wheeled mobile robot together with the real-time tip-over prevention and path following control using fuzzy and neural-fuzzy approaches can be found in \cite{b25}. 

Based on the literature review above, the present study aims to provide a solution for the identified gap between the car and mobile manipulator dynamics. Removing the tipping-over axis restrictions aims to establish a trend of monitoring supporting forces as the most relevant tipping-over stability factor in wheeled mobile manipulators.

\newpage

\section{Choosing the modelling formalism}

\label{sec:formalism}

Among the various approaches for dynamics modelling, the Lagrange formulation, based on kinetic and potential energies, and the N--E formulation, based on the balance of forces acting on a rigid manipulator link, are the most common, with the N--E approach considered as more fundamental, \cite{b26}.

In the recursive N--E algorithm (RNEA), the number of computations increases linearly with the number of degrees of freedom (DOF). Using the RNEA, linear/angular velocity vectors are calculated from a manipulator arm base to a manipulator arm tip. Forces/moments are calculated in reverse order, going from the manipulator arm tip to the arm base. In the case of mobile manipulators, it will be interesting to note that the kinematics analysis starts with the chassis and branches towards each wheel and the manipulator arm tool centre-point (TCP). Irrespectively of the underlying case, kinematic relations are the first that must be appropriately established since all the subsequent results depend on them.

Reformulations of the RNEA equations using 6D vectors where linear and angular velocities are stacked together, apart from leading to the more compact notation, let a problem be solved more directly, at a higher level of abstraction, \cite{b27}. Mathematical models formulated using the 6D vector RNEA are also indispensable, for example, in the virtual decomposition control (VDC) field of research, \cite{zhuVDC}.

Apart from the 6D RNEA benefits mentioned above, the primary motivation for using the N--E formulation here is that the free-body diagram analysis allows the direct inclusion of ground reaction forces in the dynamics analysis. In line with this direct inclusion of the manipulator supporting forces, all the relevant geometric and inertial properties are naturally included and will participate in expressions, which will be derived here as the main result.

A 6D vector dynamics model of a mobile manipulator will be derived starting from deriving a wheel dynamics model, followed by deriving a chassis dynamics model. As a manipulator arm, a serial-parallel hydraulic manipulator will be used and modelled using the state-of-the-art N--E model given in \cite{b28}, although any N--E model of any manipulator arm would fit in the proposed narrative.

\section{Mathematical preliminaries}

\label{sec: preliminaries}

Inevitable terms and notions are presented here from \cite{b29}.

Every rigid body in the analysis will have at least one three-dimensional coordinate system $\left\lbrace \bf A \right\rbrace$ (called frame $\left\lbrace \bf A \right\rbrace$ in the following text) attached to it.

Let the linear and angular velocities as sensed in frame $\left\lbrace \bf A \right\rbrace$ be denoted throughout the paper as ${^{\bf A}\boldsymbol{v}} = \begin{pmatrix}
	{^{\bf A} v_{\rm x}} & {^{\bf A} v_{\rm y}} & {^{\bf A} v_{\rm z}}
\end{pmatrix}^{T}$ and  ${^{\bf A} \boldsymbol{\omega}} = \begin{pmatrix}
	{^{\bf A} \omega_{\rm x}} & {^{\bf A} \omega_{\rm y}} & {^{\bf A} \omega_{\rm z}}
\end{pmatrix}^{T}$, respectively. Further, adopting the notation from \cite{b29}, the 6D linear/angular velocity vector in frame $\left\lbrace \bf A \right\rbrace$ is:
\begin{equation}
	{^{\bf A} \boldsymbol{V}} = \begin{pmatrix}
		{^{\bf A}\boldsymbol{v}^{T}} & {^{\bf A}\boldsymbol{\omega}^{T}}
	\end{pmatrix}^{T} \in \mathbb{R}^6.
	\label{eqn: AV}
\end{equation}

Let the force and moment vectors applied to the origin of frame $\left\lbrace \bf A \right\rbrace$  be similarly denoted as velocities using notation ${^{\bf A}\boldsymbol{f}} = \begin{pmatrix}
	{^{\bf A} f_{\rm x}} & {^{\bf A} f_{\rm y}} & {^{\bf A} f_{\rm z}}
\end{pmatrix}^{T}$ for forces and similar  notation ${^{\bf A} \boldsymbol{m}} = \begin{pmatrix}
	{^{\bf A} m_{\rm x}} & {^{\bf A} m_{\rm y}} & {^{\bf A} m_{\rm z}}
\end{pmatrix}^{T}$ for moments. Similarly to \eqref{eqn: AV}, the 6D force/moment vector, as sensed and expressed in frame $\left\lbrace \bf A \right\rbrace$, is introduced as:
\begin{equation}
	{^{\bf A} \boldsymbol{F}} = \begin{pmatrix}
		{^{\bf A}\boldsymbol{f}^{T}} & {^{\bf A}\boldsymbol{m}^{T}}
	\end{pmatrix}^{T} \in \mathbb{R}^6.
	\label{eqn: AF}
\end{equation}

Further, let frame $\left\lbrace \bf B \right\rbrace$  be attached to the same rigid body as frame $\left\lbrace \bf A \right\rbrace$. Moving the force from the frame $\left\lbrace \bf A \right\rbrace$ origin to the frame $\left\lbrace \bf B \right\rbrace$ origin introduces the moment of that force about the frame $\left\lbrace \bf B \right\rbrace$ origin. Then, quantities from (\ref{eqn: AV}) and  (\ref{eqn: AF}) can be transformed among the frames as:
\begin{equation}
	{^{\bf B} \boldsymbol{V}} = {^{\bf A} \mathbf{U}_\mathbf{B}^T} \, 	{^{\bf A} \boldsymbol{V}},
	\label{eqn: BV trans}
\end{equation}
and
\begin{equation}
	{^{\bf A} \boldsymbol{F}} = {^{\bf A}\mathbf{U_B}} \, 	{^{\bf B} \boldsymbol{F}},
	\label{eqn: BF trans}
\end{equation}
where ${^{\bf A}\mathbf{U_B}} \in \mathbb{R}^{6 \times 6}$ in  (\ref{eqn: BV trans}) and (\ref{eqn: BF trans}) is a force/moment transformation matrix, transforming the force/moment vector measured and expressed in frame $\left\lbrace \bf B \right\rbrace$ to the same force/moment vector measured and expressed in frame $\left\lbrace \bf A \right\rbrace$. The transformation matrix can be further written as:
\begin{equation}
	{^{\bf A}\mathbf{U_B}} = \begin{pmatrix}
		{^{\bf A}\mathbf{R_B}} & \mathbf{O}_{3 \times 3} \\ \left( {^{\bf A}\boldsymbol{r}_\mathbf{{AB}}} \times  \right)
		{^{\bf A}\mathbf{R_B}} & {^{\bf A}\mathbf{R_B}} \\ 
	\end{pmatrix},
	\label{eqn: AUB}
\end{equation}
where  ${^{\bf A}\mathbf{R_B}} \in \mathbb{R}^{3 \times 3}$ is a rotation (direction cosine) matrix from  frame $\left\lbrace \bf A \right\rbrace$ to frame $\left\lbrace \bf B \right\rbrace$, $\mathbf{O}_{3 \times 3}$ denotes 3-by-3 zero-matrix, and $\left(  {^{\bf A}\boldsymbol{r}_\mathbf{{AB}}}  \times  \right)$ in \eqref{eqn: AUB} is a skew-symmetric matrix operator, intended for cross-product calculation, defined as:
\begin{equation}
	\left(  {^{\bf A}\boldsymbol{r}_\mathbf{{AB}}}  \times  \right) = \begin{pmatrix}
		0 & -r_{\rm z} & r_{\rm y} \\ r_{\rm z} & 0 & -r_{\rm x} \\ -r_{\rm y} & r_{\rm x} & 0
	\end{pmatrix},
	\label{eqn: skewsymmetric operator}
\end{equation}
with $r_{\rm x}$, $r_{\rm y}$ and $r_{\rm z}$ denoting distances from the origin of frame $\left\lbrace \bf A \right\rbrace$ to the origin of frame $\left\lbrace \bf B \right\rbrace$ along the frame $\left\lbrace \bf A \right\rbrace$ $x$-, $y$- and $z$-axis, respectively. 

The net force/moment vector ${^{\bf A} \boldsymbol{F}^*} \in \mathbb{R}^{6}$ of a rigid body, in frame $\left\lbrace \bf A \right\rbrace$ is defined as:
\begin{equation}
	\mathbf{M}_{\bf A} \dfrac{\rm d}{\mathrm{d}t} \left( {^{\bf A}\boldsymbol{V}} \right) + \mathbf{C}_{\bf A} \left( {^{\bf A}\boldsymbol{\omega}}  \right) {^{\bf A}\boldsymbol{V}} + \mathbf{G}_{\bf A} = {^{\bf A} \boldsymbol{F}^*},
	\label{eqn: tot force}
\end{equation}
where $\mathbf{M}_{\bf A} \in \mathbb{R}^{6 \times 6}$ is the mass matrix, $ \mathbf{C}_{\bf A} \left( {^{\bf A}\boldsymbol{\omega}}  \right) \in \mathbb{R}^{6 \times 6}$ is the matrix of Coriolis and centrifugal terms, and  $\mathbf{G}_{\bf A} \in \mathbb{R}^{6}$ includes the gravity terms. Detailed expressions for matrices in \eqref{eqn: tot force} when a body-fixed frame adopts all the body motions are given in \cite{b29}. In cases when it is not convenient for the underlying analysis to assume that the body-fixed frame adopts all the body motions, as will be here when analysing  wheel dynamics, a straightforward application of expressions from \cite{b29} is not possible, and reformulation is needed.

\section{Kinematic chain}

\label{sec: kinematic chain}

Let us observe a chassis of a mobile manipulator  negotiating a slope in Fig. \ref{fig1}. Manipulator arm and wheels are not shown here for the sake of visibility and clarity. Their respective orientations are defined relative to the chassis orientation and will be addressed when required and appropriate.

\Figure[t!](topskip=0pt, botskip=0pt, midskip=0pt)[width=.7\textwidth]{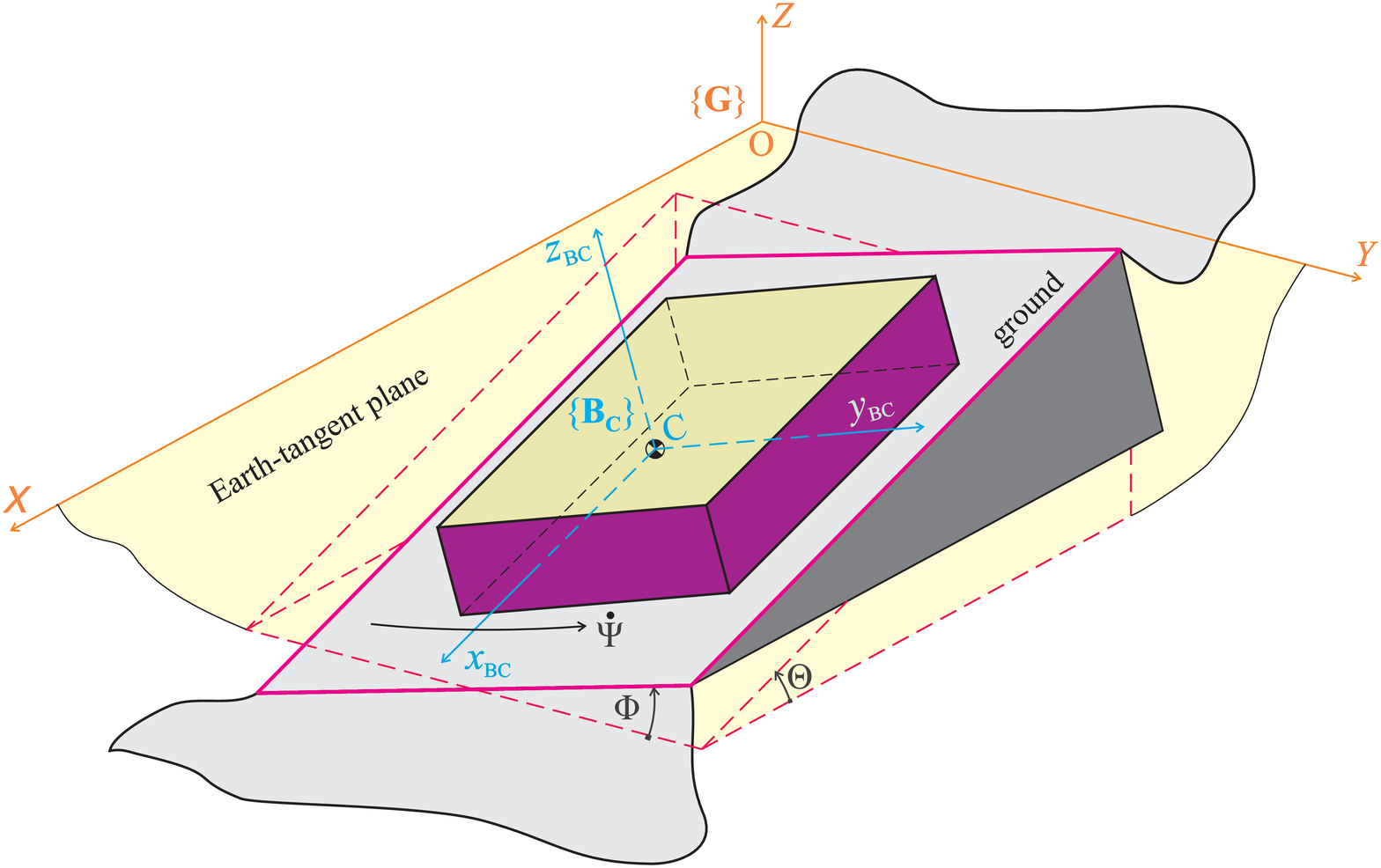}
{Manipulator chassis negotiating a slope. Uneven terrain increases the tipping-over risk and must be appropriately considered in the underlying analysis.\label{fig1}}

For the sake of clarity, it is necessary to address certain underlying assumptions before proceeding with the analysis since they will strongly affect the derivation of kinematics and dynamics relations.  

Firstly, considering that some mobile manipulators may be navigated over long distances on the Earth surface, it is worth discussing the assumption related to the inertial frame choice for the analysis completeness, as in \cite{b30}. The Earth rotation's effect is considered negligible if the body moves with a velocity less than 600 m/s. When analysing all the mobile manipulators, selecting an Earth-fixed frame as an inertial frame is more than justifiable.
\begin{assumption}
	The Earth-fixed frame can serve as an inertial reference frame.
	\label{ass: Earth-fixed}
\end{assumption}
\begin{corollary}
	The angular velocity of Earth's rotation is not crucial for further considerations. 
\end{corollary}
The navigation of a mobile manipulator, especially when considering field machines that can traverse long distances, requires addressing the justifiability of the assumption that the mobile manipulator is moving in the plane locally tangent to the Earth. Assuming that the manipulator will move in an area less than 100 km in diameter, it is reasonable to assume that disregarding the Earth's curvature will not introduce a noteworthy modelling error.
\begin{assumption}
	The vehicle's orientation in the tangent-plane coordinate system is a good approximation of the exact geographic attitude at the given position, and the height above the tangent plane is a good approximation of the elevation above the Earth's surface. The horizontal projection of the position vector gives a satisfactory approximation to the distance travelled over Earth's surface.
	\label{ass: Earth-tangent}
\end{assumption}
\begin{corollary}
	The position of a mobile manipulator is the position with respect to the fixed point on the plane, which is tangent to the Earth and is close to the exact mobile manipulator's trajectory. 
\end{corollary}
Gravity acceleration, in general, varies with position over the Earth, changing its values with latitude, longitude and elevation. In mobile manipulators, where inherently complex dynamical behaviour provides challenges, it is valid to assume the gravity field as a constant. This assumption will presumably not introduce a significant modelling error.
\begin{assumption}
	The gravity acceleration vector is perpendicular to the local Earth-tangent plane, and it is constant.
	\label{ass: gravity}
\end{assumption}
\begin{corollary}
	The position over the globe does not affect the behaviour of the mobile manipulator considered. The constant gravity acceleration will be labelled with $g$.
\end{corollary}
Assumptions \ref{ass: Earth-fixed} -- \ref{ass: gravity} explicitly justify using "flat-Earth" equations to address mobile manipulators' tipping-over stability. The same assumptions have also been implicitly used in the previously proposed studies.
\begin{assumption}
	"Flat-Earth" equations sufficiently good describe the mobile manipulator dynamics. 
	\label{ass: flat earth}
\end{assumption}
\begin{corollary}
	Description of a mobile manipulator dynamic behaviour over a small area of non-rotating Earth is sufficiently exact for the simulation and analysis needs.
\end{corollary}
Earth-fixed frame $\left\lbrace \mathbf{G} \right\rbrace$ (O$XYZ$) from Fig. \ref{fig1} will be considered an inertial frame throughout the following analysis. The body-fixed frame $\left\lbrace \mathbf{B_C} \right\rbrace$ $ (\mathrm{C}x_{\mathrm{BC}}y_{\mathrm{BC}}z_{\mathrm{BC}})$ with its origin in the chassis COM is chosen to adopt  all the linear/angular body motions.

\newpage

\begin{strip}
	\begin{equation}
		{\mathbf{^{G}R_{B_C}}} = \begin{pmatrix}
			\mathrm{cos} \Theta \, \mathrm{cos} \Psi + \mathrm{sin} \Theta \,  \mathrm{sin} \Phi \, \mathrm{sin} \Psi &  \mathrm{cos} \Psi \, \mathrm{sin} \Theta \, \mathrm{sin} \Phi - \mathrm{cos} \Theta \, \mathrm{sin} \Psi  &  \mathrm{cos} \Phi \, \mathrm{sin} \Theta \\
			\mathrm{cos} \Phi \, \mathrm{sin} \Psi & \mathrm{cos} \Phi \, \mathrm{cos} \Psi & -\mathrm{sin} \Phi  \\
			-\mathrm{cos} \Psi \, \mathrm{sin} \Theta + \mathrm{cos} \Theta \,  \mathrm{sin} \Phi \, \mathrm{sin} \Psi & \mathrm{cos} \Theta \, \mathrm{cos} \Psi \, \mathrm{sin} \Phi + \mathrm{sin} \Theta \,  \mathrm{sin} \Psi & \mathrm{cos} \Theta \,  \mathrm{cos} \Phi 
		\end{pmatrix}
		\label{eqn: GRBC}
	\end{equation}
\end{strip}
The transformation matrix relating frames $\left\lbrace \mathbf{G} \right\rbrace$ and $\left\lbrace \mathbf{B_C} \right\rbrace$ will be given by the  $ y_{\mathrm{BC}} - x_{\mathrm{BC}} - z_{\mathrm{BC}}$ rotation sequence to align the two considered frames. It is given with \eqref{eqn: GRBC}.

Angles $\Phi$ and $\Theta$ are related to the terrain slope and the $\Psi$ angle to the orientation in the local ground plane, as illustrated in Fig. \ref{fig1}. A rotation matrix that relates the body $\left\lbrace \mathbf{B_C} \right\rbrace$ and inertial frame of reference $\left\lbrace \mathbf{G} \right\rbrace$ enables easy extraction of angular velocity components in the $\left\lbrace \mathbf{B_C} \right\rbrace$ frame from the skew-symmetric matrix:
\begin{equation}
	\left({^{\mathbf{B_C}}_\mathbf{G}{{{\boldsymbol{\omega}}}}_{\mathbf{B_C}}} \times \right) = {\mathbf{^{B_C}{R}_{G}}}  \, {\mathbf{^{G}\dot{R}_{B_C}}},
	\label{eqn: angular velocity relation}
\end{equation}
which is formed per pattern in (\ref{eqn: skewsymmetric operator}), and where ${^{\mathbf{B_C}}_\mathbf{G}{{{\boldsymbol{\omega}}}}_{\mathbf{B_C}}}$ labels angular velocity of the chassis with respect to the inertial frame of reference $\left\lbrace \mathbf{G} \right\rbrace$, expressed in the $\left\lbrace \mathbf{B_C} \right\rbrace$ frame. Employing (\ref{eqn: GRBC}) in (\ref{eqn: angular velocity relation}) in combination with (\ref{eqn: skewsymmetric operator})  provides:
\begin{equation}
	{^{\mathbf{B_C}}_\mathbf{G}{{{\boldsymbol{\omega}}}}_{\mathbf{B_C}}} = \begin{pmatrix}  \dot{\Phi} \, \cos \Psi + \dot{\Theta} \, \cos \Phi \, \sin \Psi \\ - \dot{\Phi} \, \sin {\Psi} + \dot{\Theta} \, \cos \Phi \, \cos \Psi \\ \dot{\Psi} - \dot{\Theta} \, \sin \Phi   \end{pmatrix},
	\label{eqn: chassis angular velocity vector in BC}
\end{equation}
which is one of the essential factors required to be known in the following kinematics and dynamics analysis.

A simple example motivates the discussion regarding the terrain slope properties. When a wheeled platform is transiting an uneven terrain, strictly speaking, in the general case, each wheel is experiencing different ground slope values and gradients. This traversing introduces error in calculations if the local terrain slopes beneath each wheel are not obtained and accounted for correctly, as could be done using reconstruction from high-density laser scans \cite{LaserTerrain}. Not accounting for the terrain properties in detail may not be that significant, especially in the case of gentler terrains with no ruggedness. Even when the terrain is not locally placid, the introduced error may not be effective for a long time. These considerations also give rise to the idea of modifying the tip-over stability criterion to emphasise the terrain slope properties further.

\Figure[t!](topskip=0pt, botskip=0pt, midskip=0pt)[width=0.6\linewidth]{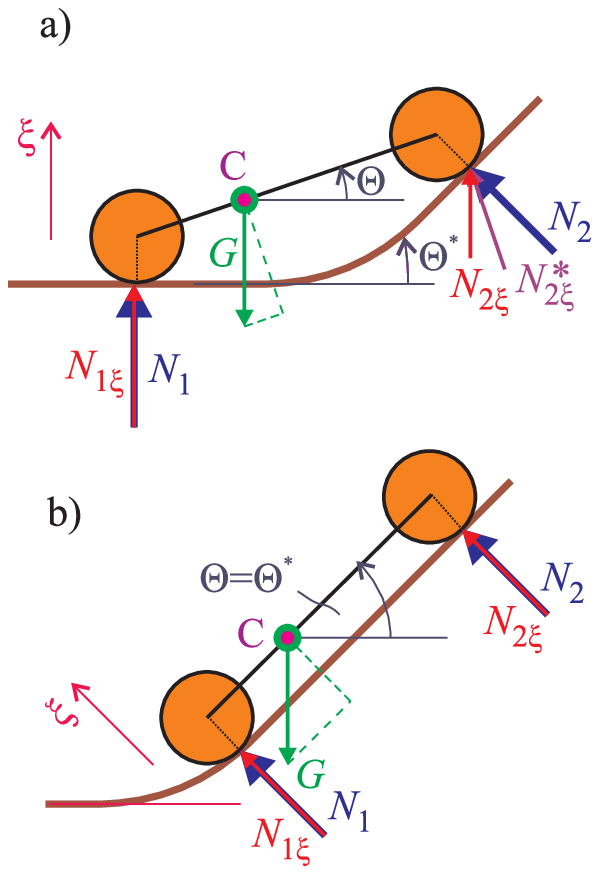}
{Manipulator wheels in different static positions. When not accounted for correctly, varying terrain slope may introduce modelling errors. \label{fig2}}

Known techniques for  IMU and GNSS measurements integration can provide chassis orientation indicators, such as Euler angles. Very often, only these values are available in on-line calculations. Let us observe the simplest possible case of two massless wheels at rest, connected with a heavy rod having the centre of mass C, as in Fig. \ref{fig2}. Without too much detailing, let us consider only equations for the sum of forces in the $\xi$-direction since this explains the idea. It is borne in mind that only normal supporting forces are of interest. In case a), expressing them explicitly and remembering the assumed static conditions, the correct equation describing the sum of forces in the  $\xi$-direction is:
\begin{equation}
	N_{1} + N_{2} \cos \Theta^* = G.
	\label{eqn: correct balance}
\end{equation}

As is argued, when the value of the slope angle $\Theta^*$ is not available, and its use during the manipulator operational phase is not possible, the best equation that can be written as a part of the on-line computation procedure would be:
\begin{equation}
	N_{1} + N_{2} \cos \Theta = G.
	\label{eqn: wrong balance}
\end{equation}

From \eqref{eqn: correct balance} and \eqref{eqn: wrong balance}, it becomes clear that an error is consciously introduced, and it depends on the difference between angles $\Theta^*$ and $\Theta$. The discussed error should be acceptable in many practical applications, excluding the highly rugged terrains with sporadic topography anomalies.  In case b), considering that $\Theta = \Theta^*$, the correct equation:
\begin{equation}
	N_1 + N_2 = G \, \cos \Theta,
	\label{eqn: correct balance locally flat}
\end{equation}
can be formed. If the considered wheels were moving from position a) to position b), it becomes clear that the inevitable model error would exist for some time, becoming eventually equal to zero. From \eqref{eqn: correct balance locally flat}, it also becomes apparent that if all four wheels were ideally in the same plane, then the estimated vehicle orientation with respect to the inertial frame of reference could perfectly capture the terrain slope and ideally enable forming of error-free equations of motion.

Even from these elementary cases, hardships of comprehending terrain properties that introduce some modelling error become evident. Based on the discussion above,  the common assumption that the terrain is locally flat and does not deform, as is sketched in Fig. \ref{fig1}, rises to the occasion.

\newpage

\begin{assumption}
	All the wheels make contact with the same, locally flat, ground plane.
	\label{ass: flat-ground}
\end{assumption}
\begin{corollary}
	 The supporting forces acting on the manipulator wheels are mutually parallel.
\end{corollary}

It should be borne in mind that angle values  ${\Phi}$ and  ${\Theta}$ are dominant and  influential in the tipping-over analysis.

Angular velocity components $\dot{\Phi}$ and  $\dot{\Theta}$ in (\ref{eqn: chassis angular velocity vector in BC}) are a simple consequence of terrain geometrical properties, and by no means these should be neglected in the orientation determination algorithms. If the traversed terrain is not challenging, assuming mild slope gradients, the contribution of terms $\dot{\Phi}$ and  $\dot{\Theta}$ in the analysis of the forces can be neglected. Even if the terrain is challenging, probably, the manipulator will not traverse it with large velocities such that these contributions in forces calculations may not be significant. 
\begin{assumption}
	The terrain itself and how the manipulator traverses uneven terrain does not significantly affect the chassis angular velocity components. 
	\label{ass: flat-ground 1}
\end{assumption}
\begin{corollary}
	Angular velocity components $\dot{\Phi}$ and $\dot{\Theta}$ are significantly smaller and less influential than the steering angular velocity $\dot{\Psi}$ and thus can be neglected, causing the following expression to hold approximately:
	\begin{equation}
		{^{\mathbf{B_C}}_\mathbf{G}{{{\boldsymbol{\omega}}}}_{\mathbf{B_C}}} = \begin{pmatrix}  0 \\ 0\\ \dot{\Psi}   \end{pmatrix}.
		\label{eqn: simplified chassis angular velocity}
	\end{equation}
\end{corollary}

Equation \eqref{eqn: simplified chassis angular velocity} follows from \eqref{eqn: chassis angular velocity vector in BC}, based on Assumption \ref{ass: flat-ground 1}. Assumptions \ref{ass: Earth-fixed} - \ref{ass: flat-ground 1} provide a basis for formulating the kinematics and dynamics relations.  In the following discussion, subscripts $\mathrm{FL}$, $\mathrm{FR}$, $\mathrm{RR}$, $\mathrm{RL}$ will be used very often and  denote the \textbf{F}ront/\textbf{R}ear \textbf{L}eft/\textbf{R}ight (usually wheel).

\begin{figure*} 
	\centering
	{\includegraphics[width=0.6\textwidth]{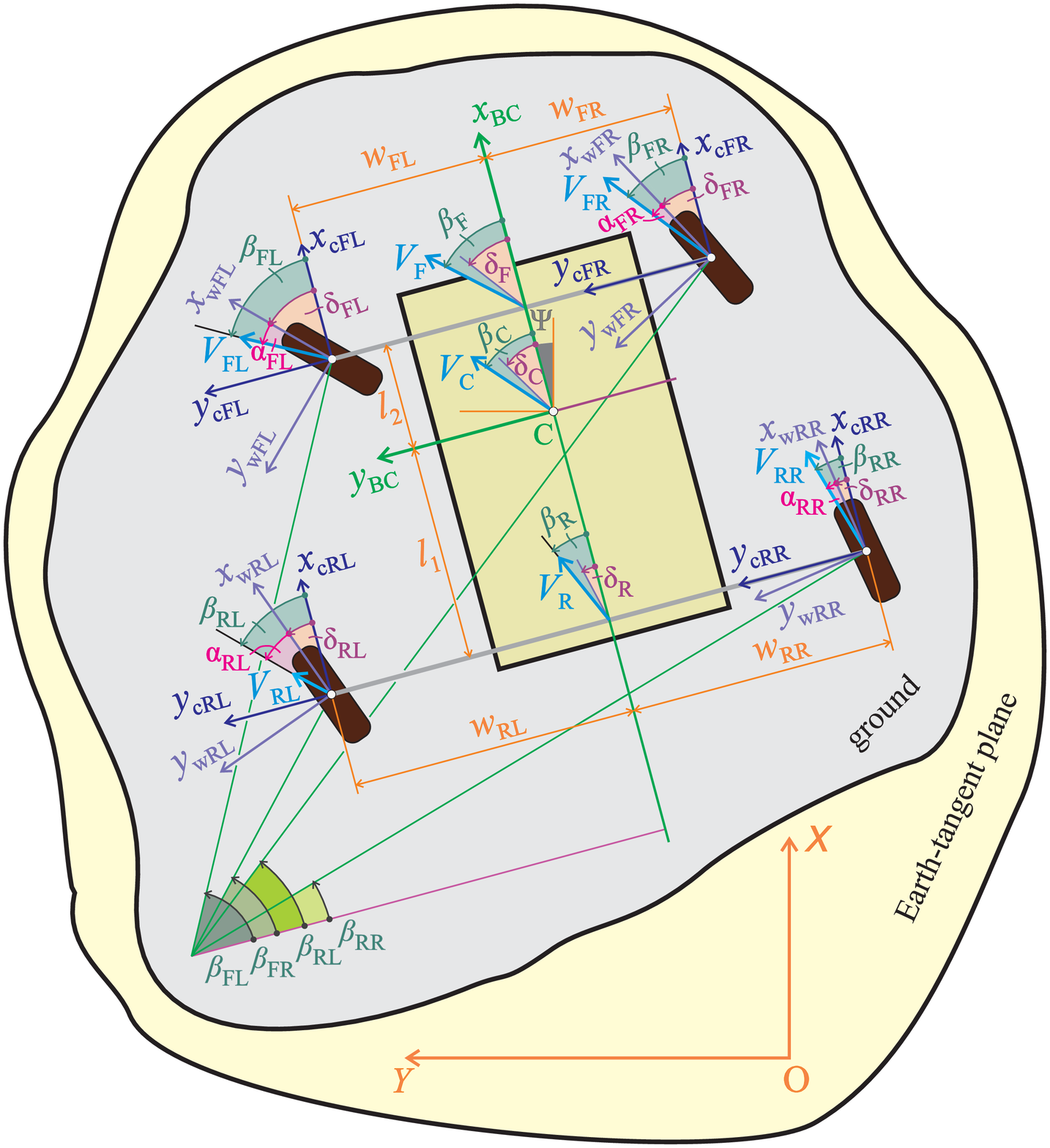}}
	\caption{A wheeled manipulator platform when negotiating a slope, with all the relevant quantities in the kinematic analysis shown.}
	\label{fig: platform kinematics}
\end{figure*}

Let us further consider the 4-AWD platform from Fig. \ref{fig: platform kinematics} which is the same manipulator wheeled platform as in Fig. \ref{fig1}, now shown in more detail, including wheels. The terms chassis and wheeled platform may be used interchangeably.

The wheeled platform's COM is located at point $\rm C$, which also acts as the origin of the right-handed frame $\left\lbrace \mathbf{B_C} \right\rbrace$ with its $x_{\mathrm{BC}}$-axis pointing in the longitudinal direction and with the $y_{\mathrm{BC}}$-axis pointing in the lateral direction of the chassis motion. The position of each wheel with respect to the COM is determined with the set of 6 fixed lengths, labelled as $l_1$, $l_2$, and $w_i$, $i=\mathrm{FL, FR, RR, RL}$. 

\Figure[t!](topskip=0pt, botskip=0pt, midskip=0pt)[width=.9\linewidth]{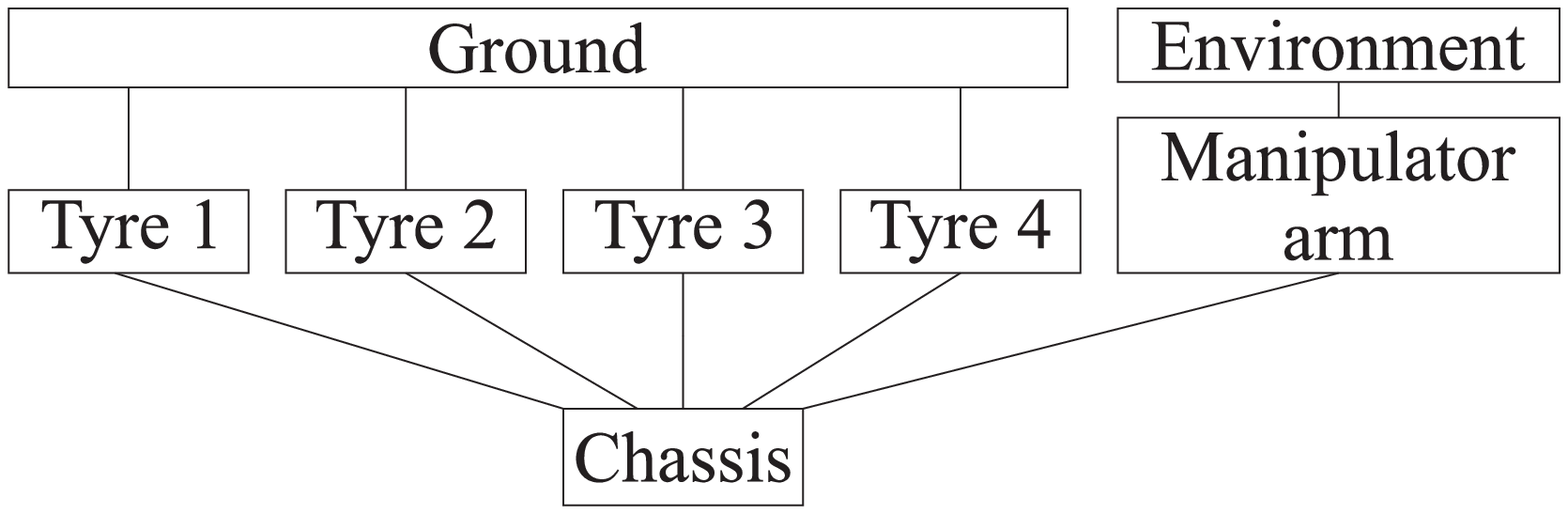}
{Simple oriented graph of the analysed mobile manipulator. \label{fig4}}

Fig. \ref{fig4} shows a mobile manipulator's simple oriented graph (SOG). From the SOG in Fig. \ref{fig4}, it can be seen that systematic kinematics calculations start from the velocity of the chassis COM and then branch towards wheels and manipulator TCP on the manipulator arm. In the considered case of the platform's planar motion, as a consequence of Assumption \ref{ass: flat-ground 1}, the chassis will presumably have velocity components only in the local level-ground plane. Each wheel's linear velocity can be calculated by knowing the magnitude and direction of the chassis linear velocity. Let the vector of the chassis COM linear velocities, with respect to the inertial frame of reference, and expressed as would be measured in the local chassis frame $\left\lbrace \mathbf{B_C} \right\rbrace$, be:
\begin{equation}
	{^{\mathbf{B_C}}_{\mathbf{G}}{{{\boldsymbol{v}_{\mathbf{B_C}}}}}} = \begin{pmatrix}
		V_{\rm C} \, \cos \beta_{\rm C} \\ V_{\rm C} \, \sin \beta_{\rm C}  \\  0
	\end{pmatrix} = \begin{pmatrix}
		{^{\mathbf{B_C}}{v}_{\mathrm{x}}} \\ {^{\mathbf{B_C}}{v}_{\mathrm{y}}}  \\  0
	\end{pmatrix},
\label{eqn: vBC}
\end{equation} 
where $V_{\rm C}$ denotes the magnitude of the chassis COM velocity, and $\beta_{\rm C}$ denotes the instantaneous velocity angle of the chassis. If the angular velocity of the chassis is determined with (\ref{eqn: simplified chassis angular velocity}), then by knowing values for ${^{\mathbf{B_C}}{v}_{\mathrm{x}}}$, ${^{\mathbf{B_C}}{v}_{\mathrm{y}}}$ and $\dot{\Psi}$, magnitudes of all the wheel-related  $V_i$,  $i=\mathrm{FL, FR, RR, RL}$ and auxiliary $ i = \mathrm{F, R}$ velocities  can be calculated as \cite{kinJazar}:
\begin{equation}
	V_{\mathrm{FL}} \, \sin \beta_{\mathrm{FL}} = V_{\rm C} \, \sin \beta_{\rm C} + l_2 \, \dot{\Psi},
	\label{eqn: VFL}
\end{equation}
\begin{equation}
	V_{\mathrm{FL}} \, \cos \beta_{\mathrm{FL}} = V_{\rm C} \, \cos \beta_{\rm C} - w_{\mathrm{FL}} \, \dot{\Psi},
\end{equation}
\begin{equation}
	V_{\mathrm{FR}} \, \sin \beta_{\mathrm{FR}} = V_{\rm C} \, \sin \beta_{\rm C} + l_2 \, \dot{\Psi},
\end{equation}
\begin{equation}
	V_{\mathrm{FR}} \, \cos \beta_{\mathrm{FR}} = V_{\rm C} \, \cos \beta_{\rm C} + w_{\mathrm{FR}} \, \dot{\Psi},
\end{equation}
\begin{equation}
	V_{\mathrm{RR}} \, \sin \beta_{\mathrm{RR}} = V_{\rm C} \, \sin \beta_{\rm C} - l_1 \, \dot{\Psi},
\end{equation}
\begin{equation}
	V_{\mathrm{RR}} \, \cos \beta_{\mathrm{RR}} = V_{\rm C} \, \cos \beta_{\rm C} + w_{\mathrm{RR}} \, \dot{\Psi},
\end{equation}
\begin{equation}
	V_{\mathrm{RL}} \, \sin \beta_{\mathrm{RL}} = V_{\rm C} \, \sin \beta_{\rm C} - l_1 \, \dot{\Psi},
\end{equation}
\begin{equation}
	V_{\mathrm{RL}} \, \cos \beta_{\mathrm{RL}} = V_{\rm C} \, \cos \beta_{\rm C} - w_{\mathrm{RL}} \, \dot{\Psi},
\end{equation}
\begin{equation}
	V_{\rm F} \, \sin \beta_{\rm F} = V_{\rm C} \, \sin \beta_{\rm C} + l_2 \, \dot{\Psi},
\end{equation}
\begin{equation}
	V_{\rm F} \, \cos \beta_{\rm F} = V_{\rm C} \, \cos \beta_{\rm C}, 
\end{equation}
\begin{equation}
	V_{\rm R} \, \sin \beta_{\rm R} = V_{\rm C} \, \sin \beta_{\rm C} - l_1 \, \dot{\Psi},
\end{equation}
\begin{equation}
	V_{\rm F} \, \cos \beta_{\rm R} = V_{\rm C} \, \cos \beta_{\rm R},
\end{equation}
where the instantaneous velocity direction angles  $\beta_i$, $i = \text{FL, FR, RR, RL, F, R}$ are determined as:
\begin{equation}
	\beta_{\mathrm{FL}} = \arctan \dfrac{V_{\rm C} \, \sin \beta_{\rm C} + l_2 \, \dot{\Psi}}{V_{\rm C} \, \cos \beta_{\rm C} - w_{\mathrm{FL}} \, \dot{\Psi}},
\end{equation}
\begin{equation}
	\beta_{\mathrm{FR}} = \arctan \dfrac{V_{\rm C} \, \sin \beta_{\rm C} + l_2 \, \dot{\Psi}}{V_{\rm C} \, \cos \beta_{\rm C} + w_{\mathrm{FR}} \, \dot{\Psi}},
\end{equation}
\begin{equation}
	\beta_{\mathrm{RR}} = \arctan \dfrac{V_{\rm C} \, \sin \beta_{\rm C} - l_1 \, \dot{\Psi}}{V_{\rm C} \, \cos \beta_{\rm C} + w_{\mathrm{RR}} \, \dot{\Psi}},
\end{equation}
\begin{equation}
	\beta_{\mathrm{RL}} = \arctan \dfrac{V_{\rm C} \, \sin \beta_{\rm C} - l_1 \, \dot{\Psi}}{V_{\rm C} \, \cos \beta_{\rm C} - w_{\mathrm{RL}} \, \dot{\Psi}},
\end{equation}
\begin{equation}
	\beta_{\rm F} = \arctan \dfrac{V_{\rm C} \, \sin \beta_{\rm C} + l_2 \, \dot{\Psi}}{V_{\rm C} \, \cos \beta_{\rm C}},
\end{equation}
\begin{equation}
	\beta_{\rm R} = \arctan \dfrac{V_{\rm C} \, \sin \beta_{\rm C} - l_1 \, \dot{\Psi}}{V_{\rm C} \, \cos \beta_{\rm C}}.
	\label{eqn: betaR}
\end{equation}	

Having all the instantaneous velocity angles $\beta_i$ enables calculation of the tyre and auxiliary sideslip angles, if the wheel and auxiliary steering angles $\delta_i$ are known, as:
\begin{equation}
	\alpha_i = \beta_i - \delta_i, \quad i = \text{FL, FR, RR, RL, F, R}.
	\label{eqn: alphai}
\end{equation}

Kinematic relations in a manipulator arm have to be established from case to case, and thus no general approach can be presented here, but it is clear that in the forming of a kinematics chain of the manipulator arm, the starting point will be the chassis, with the TCP at the end of the chain.

\section{6D vector mobile manipulator dynamics}

\label{sec: 6d dyn}

\subsection{Wheel dynamics}

The vehicle dynamics are mainly affected  by the tyre/road interaction, that is, by forces and moments generated under the tyres. These forces and moments are usually modelled by combining empirical and theoretical approaches. The appropriate planes and frames in which tyre/road interactions will be expressed and quantified are introduced and shown in Fig. \ref{fig: wheel}. The wheel-centre plane ({wcp}) contains the flat disk obtained by narrowing the tyre, while the vertical plane ({vp}) is always normal to the ground plane (gp), \cite{b31}.

The $i$-th tyre frame $\left\lbrace \mathbf{T}_{ i} \right\rbrace$, $i = \text{FL, FR, RR, RL}$, has its origin at the centre of the tyreprint. The tyreprint is assumed to be at the wheel-centre plane and the ground plane intersection. It follows the wheel's orientation, with its $z_{\mathrm{t}i}$-axis always in the vertical plane. The wheel frame $\left\lbrace \mathbf{W}_{ i} \right\rbrace$ has its origin at the wheel centre, where the COM is also assumed to be located. It will be chosen to move together with the wheel, except the spinning, since this benefits the dynamics analysis.

\Figure[t!](topskip=0pt, botskip=0pt, midskip=0pt)[width=.6\linewidth]{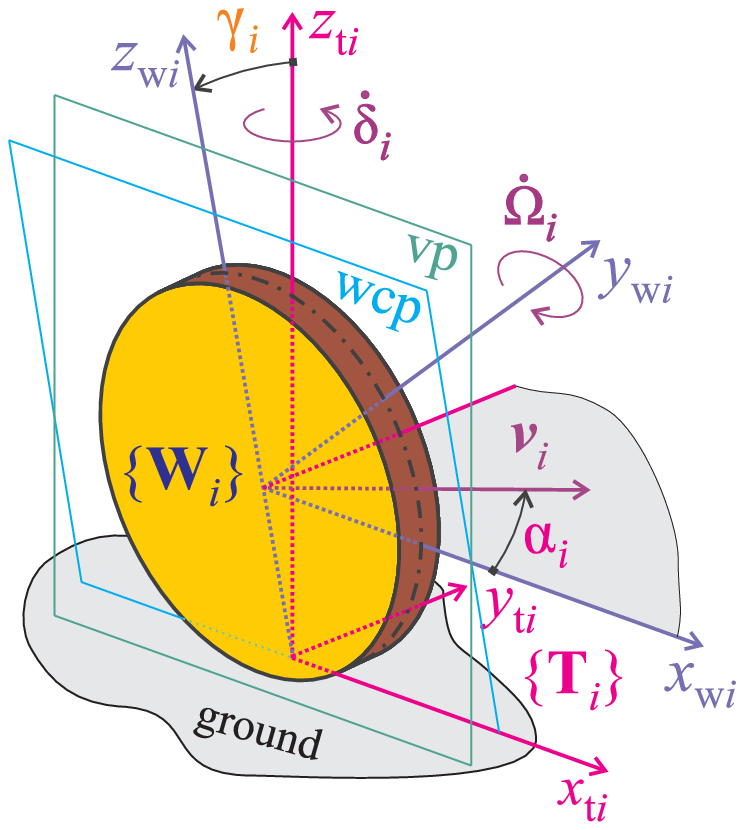}
{Wheel, with all the relevant planes, frames, axes and angles. \label{fig: wheel}}

\newpage

Although the wheel dynamics analysis benefits from using the non-spinning frame, \cite{b33},  this does not allow the straightforward use of  existing ready-to-use 6D dynamics equations as in the, for example, VDC mainstream form, where it is assumed that each body frame adopts all the body motions.

The $i$-th wheel from Fig. \ref{fig: wheel} has its COM velocity vector laying in the plane parallel to the ground plane, implying:
\begin{equation}
	{^{\mathbf{T}_i}{{{\boldsymbol{v}}}}_{i}} = \begin{pmatrix}
		V_{i} \, \cos \alpha_i \\ V_{i} \, \sin \alpha_i  \\  0
	\end{pmatrix},
\label{eqn: Tivi}
\end{equation} 
which can be expressed in the $i$-th wheel frame $\left\lbrace \mathbf{W}_i \right\rbrace$, using the sideslip angle $\alpha_i$ and the camber angle $\gamma_i$ as:
\begin{equation}
	{^{\mathbf{W}_i}{{{\boldsymbol{v}}}}_{i}} = \begin{pmatrix}
		V_{i} \, \cos \alpha_i \\ V_{i} \, \sin \alpha_i \, \cos \gamma_i \\  -V_{i} \, \sin \alpha_i \, \sin \gamma_i
	\end{pmatrix}.
\label{eqn: Wivi}
\end{equation} 

In \eqref{eqn: Tivi} and \eqref{eqn: Wivi}, kinematic relations \eqref{eqn: vBC} - \eqref{eqn: alphai} are to be consulted.
The  $i$-th wheel, considered independently from the vehicle, has the $\dot{\Omega}_i$ angular velocity component about the $y_{\mathrm{w}i}$-axis and the $\dot{\delta}_i$ component about the $z_{\mathrm{t}i}$-axis. It must be borne in mind that each wheel, considered as a part of a vehicle in the later analysis, will, in addition to these angular velocity components, adopt the angular velocity of the chassis and the term ${^{\mathbf{W}_i}_\mathbf{G}{{{\boldsymbol{\omega}}}}_{\mathbf{B_C}}}$ quantifies this, which seems to be often neglected or overlooked.
Referring to Fig. \ref{fig: wheel}, the angular velocity vector of the $i$-th wheel can be expressed in the $i$-th non-spinning wheel frame $\left\lbrace \mathbf{W}_i \right\rbrace$ as:
\begin{equation}
	{^{\mathbf{W}_i}{{{\boldsymbol{\omega}}}}_{i}} = \begin{pmatrix}
		0 \\ \dot{\Omega}_i +   \dot{\delta}_i  \, \sin \gamma_i  \\     \dot{\delta}_i   \, \cos \gamma_i
	\end{pmatrix} + {^{\mathbf{W}_i}_\mathbf{G}{{\boldsymbol{\omega}}}_{\mathbf{B_C}}}.
	\label{eqn: Wiwi}
\end{equation}

As is already mentioned, since the frame $\left\lbrace \mathbf{W}_i \right\rbrace$ is chosen not to adopt the wheel's spinning motion, quantified by $\dot{\Omega}_i$,  the angular velocity of the $i$-th wheel frame $\left\lbrace \mathbf{W}_i \right\rbrace$ \textit{per se} is:
\begin{equation}
	{^{\mathbf{W}_i}{{{\boldsymbol{\omega}}}}_{\mathbf{W}_i}} = \begin{pmatrix}
		0 \\ \dot{\delta}_i \, \sin \gamma_i \\  \dot{\delta}_i  \, \cos \gamma_i
	\end{pmatrix} + {^{\mathbf{W}_i}_\mathbf{G}{{{\boldsymbol{\omega}}}}_{\mathbf{B_C}}}.
\label{eqn: frame angular velocity}
\end{equation}

The change of linear momentum of the $i$-th wheel, expressed in the wheel frame $\left\lbrace \mathbf{W}_i \right\rbrace$, can be written as:
\begin{equation}
	m_i  {^{\mathbf{W}_i}\dot{{{\boldsymbol{v}}}}_{{i}}} +  m_i \, \left( {^{\mathbf{W}_i} {\boldsymbol{\omega}}_{\mathbf{W}_i}} \times \right) {^{\mathbf{W}_i}{{{\boldsymbol{v}}}}_{i}} +  m_i  {^{\mathbf{W}_i}{\boldsymbol{g}}} =  {^{\mathbf{W}_i}{\boldsymbol{f_i^*}}},
	\label{eqn: ch lin mom F}
\end{equation}
with $m_i$ denoting the mass of the $i$-th wheel,  ${^{\mathbf{W}_i}{\boldsymbol{g}}} = \begin{pmatrix}
	0 & 0 & g
\end{pmatrix}^T$ being the gravity acceleration vector, and ${^{\mathbf{W}_i}{\boldsymbol{f_i^*}}}$ representing the total force vector acting on the $i$-th wheel, both expressed as would be measured in the $\left\lbrace \mathbf{W}_i \right\rbrace$ frame. The remaining equations required to describe the wheel motion give the angular momentum change as:
\begin{equation}
	\left[ ^{\mathbf{W}_i}\mathbf{I} \right]   {^{\mathbf{W}_i} \dot{\boldsymbol{\omega}}_{i}} + \left( {^{\mathbf{W}_i} {\boldsymbol{\omega}}_{\mathbf{W}_i}} \times \right)  \left( \left[ ^{\mathbf{W}_i}\mathbf{I} \right]  {^{\mathbf{W}_i} {\boldsymbol{\omega}}_{i}} \right) =  {^{\mathbf{W}_i}{\boldsymbol{m_i^*}}},
	\label{eqn: angular momentum change wheel}
\end{equation}
with $\left[ ^{\mathbf{W}_i}\mathbf{I} \right]$ representing the inertia tensor of the $i$-th wheel about the $\left\lbrace \mathbf{W}_i \right\rbrace$ frame axes  and  ${^{\mathbf{W}_i}{\boldsymbol{m_i^*}}}$ representing the total external moment vector acting on the $i$-th wheel, expressed as would be measured in the $\left\lbrace \mathbf{W}_i \right\rbrace$ frame. It is noted that in both \eqref{eqn: ch lin mom F} and \eqref{eqn: angular momentum change wheel}  exists the frame angular velocity given per \eqref{eqn: frame angular velocity}, whereas \eqref{eqn: Wiwi} participates only in \eqref{eqn: angular momentum change wheel}.

At this point, the 6D dynamics model can be formed. The translational and rotational $i$-th wheel dynamics are thus described per \eqref{eqn: ch lin mom F} and \eqref{eqn: angular momentum change wheel}. By introducing the 6D linear/angular velocity vector ${^{\mathbf{W}_i}\boldsymbol{V}}_i$ per \eqref{eqn: Wivi} and \eqref{eqn: Wiwi} as:
\begin{equation}
	{^{\mathbf{W}_i}\boldsymbol{V}}_i = \begin{pmatrix}
		{^{\mathbf{W}_i}{\boldsymbol{v}}_{i}} \\ {^{\mathbf{W}_i} {\boldsymbol{\omega}}_{i}} 
	\end{pmatrix},
\end{equation}
the 6D wheel/tyre dynamics equations get the compact  form:
\begin{equation}
	{\mathbf{M}}_{\mathbf{W}_i}   {^{\mathbf{W}_i}\dot{\boldsymbol{V}}_i} + {\mathbf{C}}_{\mathbf{W}_i}   {^{\mathbf{W}_i}{\boldsymbol{V}_i}} + {\mathbf{G}}_{\mathbf{W}_i} = {^{\mathbf{W}_i}{\boldsymbol{F}}}^*,
	\label{eqn: wheel dynamics}
\end{equation}
with matrices ${\mathbf{M}}_{\mathbf{W}_i}$,  ${\mathbf{C}}_{\mathbf{W}_i}$ and the vector ${\mathbf{G}}_{\mathbf{W}_i}$ being:
\begin{equation}
	{\mathbf{M}}_{\mathbf{W}_i} = \begin{pmatrix}
		m_{i} \, \mathbf{I}_{3 \times 3} & \mathbf{O}_{3 \times 3} \\ \mathbf{O}_{3 \times 3} &  \left[ ^{\mathbf{W}_i}\mathbf{I} \right]
	\end{pmatrix},
\end{equation}
\begin{equation}
	{\mathbf{C}}_{\mathbf{W}_i} = \begin{pmatrix} m_i \,
		\left( {^{\mathbf{W}_i} {\boldsymbol{\omega}}_{\mathbf{W}_i}} \times \right) &\mathbf{O}_{3 \times 3}  \\ \mathbf{O}_{3 \times 3}  &  \left( {^{\mathbf{W}_i} {\boldsymbol{\omega}}_{\mathbf{W}_i}} \times \right) \, \left[ ^{\mathbf{W}_i}\mathbf{I} \right]
	\end{pmatrix},
	\label{eqn: matrix Cwi wheel}
\end{equation}
\begin{equation}
	{\mathbf{G}}_{\mathbf{W}_i} = \begin{pmatrix} m_{i} \,  ^{\mathbf{W}_i}\boldsymbol{g} \\  \mathbf{0}_{3 \times 1}
	\end{pmatrix}.
\end{equation}

The inertia matrix ${\mathbf{M}}_{\mathbf{W}_i}$  remains symmetric here, as when the body frame adopts all the body motions. A possibly important thing to note is that  matrix ${\mathbf{C}}_{\mathbf{W}_i}$ is now not anti-symmetric in general. 

\subsection{Wheel/chassis interaction}

The following topic requiring attention is the interaction between the wheel and chassis per the modelling modularity property and appending on the wheel/tyre subsystem dynamics analysis.
If the aerodynamic forces acting on the wheel itself are neglected, the essential and unavoidable external forces/moments to be accounted for will undoubtedly remain the tyre/road interaction forces/moments ${^{\mathbf{T}_i}{\boldsymbol{F}}}$, the wheel actuation moments ${^{\mathbf{W}_i}{\boldsymbol{F}_a}}$ and the wheel/chassis interaction forces/moments ${^{\mathbf{W}_i}{\boldsymbol{\eta}_i}}$. 

\begin{figure*}[h!]
	\centering
	\includegraphics[width=\textwidth]{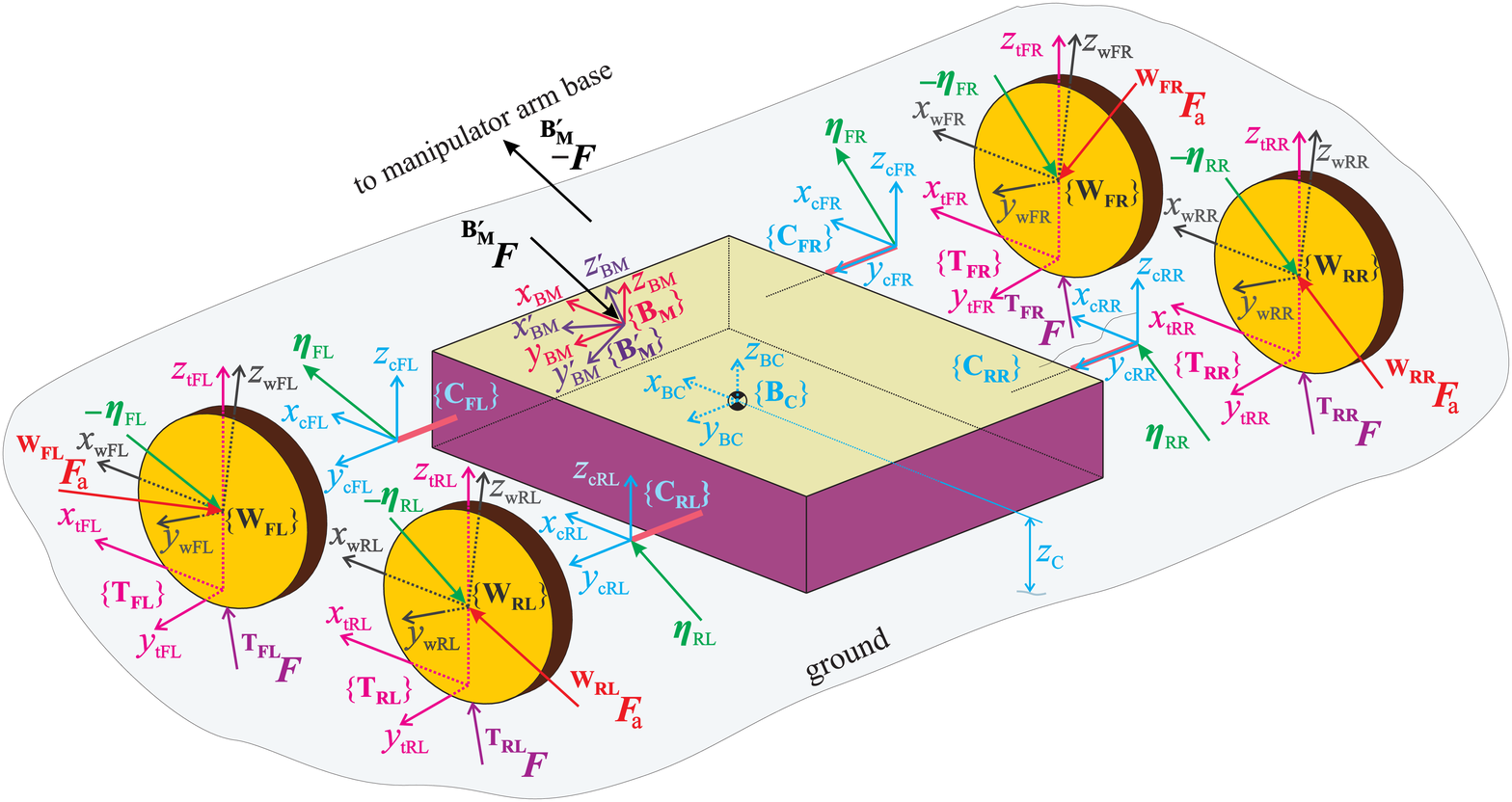}
	\caption{Chassis and wheels free body diagram. Ground reactions are smoothly included in the analysis when using the N--E formalism.} \label{fig: vehicle and wheels frames}
\end{figure*}

Referring to Fig. \ref{fig: vehicle and wheels frames} for providing graphical insight, the total force acting on each wheel, i.e. the right-hand side of  \eqref{eqn: wheel dynamics}, can be written as the following sum:
\begin{equation}
	{^{\mathbf{W}_i}{\boldsymbol{F}}}^*  =   -{^{\mathbf{W}_i}{\boldsymbol{\eta}_i}} +	{^{\mathbf{W}_i}\mathbf{U}_{\mathbf{T}_i}}{^{\mathbf{T}_i}{\boldsymbol{F}}}  + {^{\mathbf{W}_i}{\boldsymbol{F}_a}},
	\label{eqn: total wheel force}
\end{equation}
with $i = \mathrm{FL, FR, RR, RL}$ in the case considered here. Once the wheel inertial forces and the forces formed at the tyreprint centre are known, the wheel/chassis interaction force vector ${^{\mathbf{W}_i}{\boldsymbol{\eta}_i}}$, which will directly actuate the chassis, can be calculated as:
\begin{equation}
	{^{\mathbf{W}_i}{\boldsymbol{\eta}_i}}	 =  -{^{\mathbf{W}_i}{\boldsymbol{F}}}^*  +	{^{\mathbf{W}_i}\mathbf{U}_{\mathbf{T}_i}}{^{\mathbf{T}_i}{\boldsymbol{F}}}  + {^{\mathbf{W}_i}{\boldsymbol{F}_a}},
	\label{eqn: total wheel-chassis force}
\end{equation}
with $i = \mathrm{FL, FR, RR, RL}$.

Even in the general case with $n$ wheels, each wheel's dynamics could be given with \eqref{eqn: wheel dynamics}, where the total force acting on each wheel is \eqref{eqn: total wheel force}. Beginning the analysis from a manipulator with $n$ wheels and with articulated steering would undoubtedly cause the derivation of all the underlying expressions to be filled with numerous terms that could divert attention from the essence.

\newpage

\subsection{Chassis dynamics}

The focus is further turned to the chassis of the considered 4-wheel AWS vehicle in Fig. \ref{fig: vehicle and wheels frames}. The chassis dynamics, in this case, can be easily described, assuming that the frame $\left\lbrace \mathbf{B_C} \right\rbrace$ adopts all the chassis motions. 

The total force acting on the chassis is:
\begin{equation}
	\mathbf{M}_{\mathbf{B_C}}  \, {^{\mathbf{B_C}}\dot{\boldsymbol{V}}} + \mathbf{C}_{\mathbf{B_C}} \, {^{\mathbf{B_C}}{\boldsymbol{V}}} + \mathbf{G}_{\mathbf{B_C}} = {^{\mathbf{B_C}}{\boldsymbol{F}}}^*,
	\label{eqn: chassis dynamics}
\end{equation}
where the detailed expressions for matrices $\mathbf{M}_{\mathbf{B_C}}$, $\mathbf{C}_{\mathbf{B_C}}$ and $\mathbf{G}_{\mathbf{B_C}}$  are the same as in \cite{b29}. The total force acting on the chassis can be equivalently represented by the following sum: 
\begin{equation}
	{^{\mathbf{B_C}}{\boldsymbol{F}}}^* = {^{\mathbf{B_C}}\mathbf{U}_{\mathbf{B_M}}} {^{\mathbf{B_M}}{\boldsymbol{F}}} + \displaystyle\sum_{i = \mathrm{FL,..., RL}}{^{\mathbf{B_C}}\mathbf{U}_{\mathbf{W}_i}} {^{\mathbf{W}_i}{\boldsymbol{\eta}_i}}, 
	\label{eqn: total chassis force}
\end{equation}
where the ${^{\mathbf{B_M}}{\boldsymbol{F}}}$ represents the manipulator base reaction force, and the ${^{\mathbf{W}_i}{\boldsymbol{\eta}}}$ represents the wheel/chassis interaction expressed as would be measured in  $\left\lbrace \mathbf{W}_i \right\rbrace$ frame. All the existing tipping-over criteria have been derived assuming  that in (\ref{eqn: total wheel-chassis force}) terms ${^{\mathbf{W}_i}{\boldsymbol{F}}}^*$ and ${^{\mathbf{W}_i}{\boldsymbol{F}_a}}$ are equal to zero. While the MHS assumed that the left-hand side of (\ref{eqn: chassis dynamics}) equals zero, the TMO considers only the $\mathbf{G}_{\mathbf{B_C}}$ term, and the ITMO uses the complete left-hand side of (\ref{eqn: chassis dynamics}) as given here.

\section{Weight transfer to  manipulator wheels}

\label{sec: solution}

In the derivation process of MHS, TOM and ITOM, at this modelling stage, with underlying assumptions being valid, moments about the axes connecting supporting polygon vertices have been formed, and no further steps were taken towards the analytical determination of supporting forces.

As a consequence of Assumption \ref{ass: flat-ground}, the wheel supporting forces affect only the linear motion in the direction of their action and angular motions about axes in the plane parallel to the level-ground plane, i.e. pitching and rolling angular motion. Consequently, when having only three equations, the challenge in the forces determination process is that there are fewer equations than unknown variables in the case of four or more wheels. It is not seldom that additional reasonable relations for connecting the tyre loads are being introduced to obtain the closed-form system of equations, as in \cite{b34}, or the solution is found using a matrix pseudoinverse.

Before proceeding further, including the wheel forces and moments requires addressing since it may simplify the following analysis. When performing dynamics analysis at the wheel level, the wheel's inertial forces are critical in predicting the wheel's motion. On the other hand, neglecting the wheel inertial forces could lead to much simpler final expressions without introducing significant modelling errors when calculating normal loads. Let us take, for example, a moving mobile manipulator which accelerates. If the chassis mass is a few times the mass of the wheels, then its contribution to the system inertial forces is dominant. In the case of a manipulator moving at low to zero velocities during the working tasks, even the chassis inertial forces are certainly negligible since accelerations of the manipulator arm are the ones having significant velocities and accelerations. These considerations motivate the neglection of the wheel inertial forces for each wheel. If this neglection may not be valid in the particular case, the full-dynamics model presented here extends results to any desired extent. Accounting for the weight of wheels will not make the normal force expressions significantly complex and could be a significant factor, especially when considering big heavy wheels of heavy-duty machines at low velocities. 
\begin{assumption}
	Inertial forces of all the wheels are negligible with respect to the chassis inertial forces.
	\label{ass: wheel inertial forces}
\end{assumption}
\begin{corollary}
Inertial forces of each wheel and all the wheel actuation torques are neglected, leaving only respective wheel weights to be considered when calculating normal loads. In the normal load analysis, the wheel/chassis interaction force will have the following form:
	\begin{eqnarray}
		{^{\mathbf{W}_i}{\boldsymbol{\eta}_i}}	 =  -{\mathbf{G}}_{\mathbf{W}_i}  +	{^{\mathbf{W}_i}\mathbf{U}_{\mathbf{T}_i}}{^{\mathbf{T}_i}\boldsymbol{F}},
		\label{eqn: total wheel-chassis force 1}
	\end{eqnarray}
	with ${^{\mathbf{T}_i}\boldsymbol{F}} = \begin{pmatrix}
		{^{\mathbf{T}_i}{f}_x} & {^{\mathbf{T}_i}{f}_y} & {^{\mathbf{T}_i}{f}_z} & 0 & 0 & 0 
	\end{pmatrix}^T$.
\end{corollary}
In order to avoid more complicated notation in the following expressions, additional reasonable assumptions on the wheels, which are commonly understood in these kinds of analyses, are applied here. If some of these may not hold in a specific case, the more general result can be obtained following the presented procedure. Here, it is assumed that:
\begin{assumption}
	All the wheels are nominally equal, having zero camber angles, same masses, same radii, and same lateral distances from the chassis COM.
	\label{ass: same wheels}
\end{assumption}
\begin{corollary}
	All the wheel distances from the chassis $w_i$, wheel masses $m_{\mathrm{w}i}$ and wheel radii $R_{\mathrm{w}i}$ have equal values, labelled as $w$, $m_{\mathrm{w}}$ and $R_{\mathrm{w}}$, respectively, with camber angles $\gamma_i = 0$, where $i = \mathrm{FL, FR, RR, RL}$.  
\end{corollary}

\begin{strip}
	\begin{equation}
		\begin{array}{ll}
			\mathbf{z}_{f} \,	\left( {^{\mathbf{B_C}}\mathbf{U}_{\mathbf{B_M}}} {^{\mathbf{B_M}}{\boldsymbol{F}}} + \displaystyle\sum_{i = \mathrm{FL, FR, RR, RL}}{^{\mathbf{B_C}}\mathbf{U}_{\mathbf{W}_i}} {^{\mathbf{W}_i}{\boldsymbol{\eta}_i}} \right) & = \mathbf{z}_{f} \,	\left( \mathbf{M}_{\mathbf{B_C}}  \, {^{\mathbf{B_C}}\dot{\boldsymbol{V}}} + \mathbf{C}_{\mathbf{B_C}} \, {^{\mathbf{B_C}}{\boldsymbol{V}}} + \mathbf{G}_{\mathbf{B_C}} \right) \\
			&  = m_{\mathrm{BC}} \, g
		\end{array}
		\label{eqn: sum vert forces}
	\end{equation}
	\begin{equation}
		\begin{array}{ll}
			\mathbf{x}_{\tau} \, \left( {^{\mathbf{B_C}}\mathbf{U}_{\mathbf{B_M}}} {^{\mathbf{B_M}}{\boldsymbol{F}}} + \displaystyle\sum_{i = \mathrm{FL, FR, RR, RL}}{^{\mathbf{B_C}}\mathbf{U}_{\mathbf{W}_i}} {^{\mathbf{W}_i}{\boldsymbol{\eta}_i}} \right) & = \mathbf{x}_{\tau} \,	\left( \mathbf{M}_{\mathbf{B_C}}  \, {^{\mathbf{B_C}}\dot{\boldsymbol{V}}} + \mathbf{C}_{\mathbf{B_C}} \, {^{\mathbf{B_C}}{\boldsymbol{V}}} + \mathbf{G}_{\mathbf{B_C}} \right) \\
			& = -^{\mathbf{B_C}}I_{\mathrm{xy}} \, \dot{\Psi}^2 + {^{\mathbf{B_C}}I_{\mathrm{xz}}} \, \ddot{\Psi}
		\end{array}
		\label{eqn: rolling angle}
	\end{equation}
	\begin{equation}
		\begin{array}{ll}
			\mathbf{y}_{\tau} \,	\left( {^{\mathbf{B_C}}\mathbf{U}_{\mathbf{B_M}}} {^{\mathbf{B_M}}{\boldsymbol{F}}} + \displaystyle\sum_{i = \mathrm{FL, FR, RR, RL}}{^{\mathbf{B_C}}\mathbf{U}_{\mathbf{W}_i}} {^{\mathbf{W}_i}{\boldsymbol{\eta}_i}} \right) & = \mathbf{y}_{\tau} \,	\left( \mathbf{M}_{\mathbf{B_C}}  \, {^{\mathbf{B_C}}\dot{\boldsymbol{V}}} + \mathbf{C}_{\mathbf{B_C}} \, {^{\mathbf{B_C}}{\boldsymbol{V}}} + \mathbf{G}_{\mathbf{B_C}} \right) \\
			& = -^{\mathbf{B_C}}I_{\mathrm{xz}} \, \dot{\Psi}^2 + {^{\mathbf{B_C}}I_{\mathrm{yz}}} \, \ddot{\Psi}
		\end{array}
		\label{eqn: pitching angle}
	\end{equation}
\begin{equation}
	\begin{array}{ll}
		\mathbf{x}_{f} \,	\left( {^{\mathbf{B_C}}\mathbf{U}_{\mathbf{B_M}}} {^{\mathbf{B_M}}{\boldsymbol{F}}} + \displaystyle\sum_{i = \mathrm{FL, FR, RR, RL}}{^{\mathbf{B_C}}\mathbf{U}_{\mathbf{W}_i}} {^{\mathbf{W}_i}{\boldsymbol{\eta}_i}} \right)  & =  	\mathbf{x}_{f} \,	\left( \mathbf{M}_{\mathbf{B_C}}  \, {^{\mathbf{B_C}}\dot{\boldsymbol{V}}} + \mathbf{C}_{\mathbf{B_C}} \, {^{\mathbf{B_C}}{\boldsymbol{V}}} + \mathbf{G}_{\mathbf{B_C}} \right)
		\\ & = m_{\mathrm{BC}} \, \left( {^{\mathbf{B_C}}\dot{v}_{\mathrm{y}}} + {^{\mathbf{B_C}}{v}_{\mathrm{x}}} \, \dot{\Psi}  \right)
		\label{eqn: sum x forces}
	\end{array}
\end{equation}
\begin{equation}
	\begin{array}{ll}
		\mathbf{y}_{f} \,	\left( {^{\mathbf{B_C}}\mathbf{U}_{\mathbf{B_M}}} {^{\mathbf{B_M}}{\boldsymbol{F}}} + \displaystyle\sum_{i = \mathrm{FL, FR, RR, RL}}{^{\mathbf{B_C}}\mathbf{U}_{\mathbf{W}_i}} {^{\mathbf{W}_i}{\boldsymbol{\eta}_i}} \right) & = \mathbf{y}_{f} \,	\left( \mathbf{M}_{\mathbf{B_C}}  \, {^{\mathbf{B_C}}\dot{\boldsymbol{V}}} + \mathbf{C}_{\mathbf{B_C}} \, {^{\mathbf{B_C}}{\boldsymbol{V}}} + \mathbf{G}_{\mathbf{B_C}} \right) \\  & = m_{\rm{BC}} \, \left( {^{\mathbf{B_C}}\dot{v}_{\mathrm{x}}} - {^{\mathbf{B_C}}{v}_{\mathrm{y}}} \, \dot{\Psi}  \right)
		\label{eqn: sum y forces}
	\end{array}	
\end{equation}
\end{strip}
The three equations of motion in which normal loads participate follow from combination of \eqref{eqn: chassis dynamics} and \eqref{eqn: total chassis force} and these are \eqref{eqn: sum vert forces} -- \eqref{eqn: pitching angle}. Combining them with \eqref{eqn: sum x forces} and \eqref{eqn: sum y forces}
provides three equations with four unknown normal loads ${^{\mathbf{T_{FL}}}f_z}$, ${^{\mathbf{T_{FR}}}f_z}$, ${^{\mathbf{T_{RR}}}f_z}$, ${^{\mathbf{T_{RL}}}f_z}$. Auxiliary vectors used in  \eqref{eqn: sum vert forces} -- \eqref{eqn: sum y forces} are $\mathbf{x}_{f} = \begin{pmatrix}
	1&0&0&0&0&0
\end{pmatrix}^T$, $\mathbf{y}_{f} = \begin{pmatrix}
0&1&0&0&0&0
\end{pmatrix}^T$, $\mathbf{z}_{f} = \begin{pmatrix}
	0&0&1&0&0&0
\end{pmatrix}^T$, $\mathbf{x}_{\tau} = \begin{pmatrix}
0&0&0&1&0&0
\end{pmatrix}^T$, $\mathbf{y}_{\tau} = \begin{pmatrix}
0&0&0&0&1&0
\end{pmatrix}^T$.

Before presenting the main result, it is vital to address how the base frame of the manipulator arm  is oriented to the frame $\left\lbrace \mathbf{B_C} \right\rbrace$ at the chassis COM. The mutual orientation of these frames will, of course, affect the final expressions for normal wheel loads. In the general case, this orientation can be any, and if it were to be accounted for directly, many additional terms in final expressions coming from the rotation matrix coefficients would appear. For the sake of analysis clarity, it is convenient to have both the manipulator arm base $\left\lbrace \mathbf{B_M'} \right\rbrace$ frame, which adopts all the manipulator base motions, and frame $\left\lbrace \mathbf{B_M} \right\rbrace$ with the same origin as $\left\lbrace \mathbf{B_M'} \right\rbrace$, but with the same orientation as $\left\lbrace \mathbf{B_C} \right\rbrace$, per Fig. \ref{fig: vehicle and wheels frames}. By doing this, the model generality is kept as high as possible, with at the same time,  ready-to-use expressions for normal loads are derived and presented as simple as possible. The considerations above that are introduced simply for the  presentation convenience lead to the  assumption:

\newpage

\begin{assumption}
	Frame $\left\lbrace \mathbf{B_M} \right\rbrace$ at the connection of the manipulator arm base and the chassis has the same orientation as the frame $\left\lbrace \mathbf{B_C} \right\rbrace$ at the chassis COM.
	\label{ass: BC Rot BM}
\end{assumption}
\begin{corollary}
	The rotation matrix from frame $\left\lbrace \mathbf{B_C} \right\rbrace$ to the frame $\left\lbrace \mathbf{B_M} \right\rbrace$ is an identity matrix ${\mathbf{^{B_M}R_{B_C}}} = \mathbf{I}_{3 \times 3}$.
\end{corollary}

The main implication of  Assumption \ref{ass: BC Rot BM} is that an inevitable and insignificant burden is placed on the end-user to calculate ${^{\mathbf{B_M}}{\boldsymbol{F}}}$ before using the main results. If the below derived equations are to be used directly, one must transform the forces/moments from  frame $\left\lbrace \mathbf{B_M'} \right\rbrace$ to frame $\left\lbrace \mathbf{B_M} \right\rbrace$ as:
\begin{eqnarray}
	{^{\mathbf{B_M}}{\boldsymbol{F}}} = {^{\mathbf{B_M}}\mathbf{U}_{\mathbf{B_M'}}} \, {^{\mathbf{B_M'}}{\boldsymbol{F}}}.
\end{eqnarray}

Forces/moments vector ${^{\mathbf{B_M'}}{\boldsymbol{F}}}$ at the manipulator arm base can be calculated as, for example, in \cite{b28} and ${^{\mathbf{B_M}}{\boldsymbol{F}}}$ is easily obtained knowing the orientation of the manipulator arm base frame $\left\lbrace \mathbf{B_M'} \right\rbrace$ with respect to the chassis-fixed frame $\left\lbrace \mathbf{B_M} \right\rbrace$.

If the solution obtained using the pseudoinversion is to be avoided, a unique solution for the supporting forces can be obtained by introducing a suspension-related assumption. It connects the normal load differences on the front and rear axle. Following the vehicle dynamics modelling principles: 
\begin{assumption}
	\cite{b34} The lateral load difference across the front axle is some fraction of the total lateral load difference.
\end{assumption}
\begin{corollary}
	The following equation is recognized:
	\begin{equation}
		\begin{array}{l}
			{^{\mathbf{T_{FR}}}f_z} - {^{\mathbf{T_{FL}}}f_z} =    \\   \qquad D \, \left( {^{\mathbf{T_{FR}}}f_z} + {^{\mathbf{T_{RR}}}f_z} - {^{\mathbf{T_{FL}}}f_z} - {^{\mathbf{T_{RL}}}f_z} \right),
		\end{array}
		\label{eqn: normal load 4th eq}
	\end{equation}
	with $D \in [0,1]$.  
\end{corollary}
The final system of equations for the normal load calculation with $D = 0.5$ in (\ref{eqn: normal load 4th eq}) is:
\begin{equation}
	\begin{pmatrix}
		w & -w & -w & w \\ -l_{2} & -l_{2} & l_{1} & l_{1} \\ 1 & 1 & 1 & 1 \\ -0.5 & 0.5 & -0.5 & 0.5		
	\end{pmatrix}  \begin{pmatrix} {^{\mathbf{T_{FL}}}f_z} \\ {^{\mathbf{T_{FR}}}f_z} \\ {^{\mathbf{T_{RR}}}f_z} \\ {^{\mathbf{T_{RL}}}f_z} \end{pmatrix}= \begin{pmatrix}
		b_1 \\ b_2 \\ b_3	 \\ 0		
	\end{pmatrix},
	\label{eq: normal forces final eq}
\end{equation}
where the corresponding $b_i$ from \eqref{eq: normal forces final eq}, $i = 1,2,3$ are:
\begin{strip}
\begin{equation}
	\begin{array}{ll}
		b_1   = & \left(z_{\mathrm{CA}} + z_{\mathrm{C}} \right) \, {^{\mathbf{B_M}}{f}_{\mathrm{y}}} - y_{\mathrm{CA}} \, {^{\mathbf{B_M}}{f}_{\mathrm{z}}} -   {^{\mathbf{B_M}}{m}_{\mathrm{x}}} -  g \, ( m_{\mathrm{BC}} \, z_{\rm C} + 4 \, m_{\rm w} \, R_{\rm w}) \, r_{32} - \\ &   m_{\mathrm{BC}} \, z_{\mathrm{C}} \, \left( {^{\mathbf{B_C}}\dot{v}_{\mathrm{y}}} +  {^{\mathbf{B_C}}{v}_{\mathrm{x}}} \, \dot{\Psi}  \right) - {^{\mathbf{B_C}}I_{\mathrm{yz}}} \, \dot{\Psi}^2 + {^{\mathbf{B_C}}I_{\mathrm{xz}}} \, \ddot{\Psi},
	\end{array}
	\label{eq: b1}
\end{equation}
\begin{equation}
	\begin{array}{ll}
		b_2 = &  - \left(z_{\mathrm{CA}} + z_{\mathrm{C}} \right) \, {^{\mathbf{B_M}}{f}_{\mathrm{x}}} + x_{\mathrm{CA}}  \, {^{\mathbf{B_M}}{f}_{\mathrm{z}}} -  {^{\mathbf{B_M}}{m}_{\mathrm{y}}} + g \, (m_{\mathrm{BC}}  \, z_{\rm C} + 4  \, m_{\rm w} \, R_{\rm w})\, r_{31} + \\ &  2 \, g \, (l_1 - l_2) \, m_{\rm w} \, r_{33}    + m_{\mathrm{BC}} \, z_{\rm C} \, \left( {^{\mathbf{B_C}}\dot{v}_{\mathrm{x}}} - {^{\mathbf{B_C}}{v}_{\mathrm{y}}} \, \dot{\Psi}  \right) + {^{\mathbf{B_C}}I_{\mathrm{xz}}} \, \dot{\Psi}^2 + {^{\mathbf{B_C}}I_{\mathrm{yz}}} \, \ddot{\Psi},
	\end{array}
	\label{eq: b2}
\end{equation}
\begin{equation}
	b_3 =  g  \, \left( m_{\mathrm{BC}} + 4 \, m_{\mathrm{w}} \right) \, r_{33} - {^{\mathbf{B_M}}{f}_{\mathrm{z}}}.
	\label{eq: b3}
\end{equation}
\end{strip}
Particular terms in \eqref{eq: b1}--\eqref{eq: b3} with the left superscript $\mathbf{B_M}$ arise from the existence of the forces/moments being built up at the connection of the chassis and manipulator base. The vector ${^{\mathbf{B_M}}{\boldsymbol{F}}}$ is assumed to have the following structure:
\begin{equation}
	{^{\mathbf{B_M}}{\boldsymbol{F}}}=\begin{pmatrix}
		{^{\mathbf{B_M}}{f}_{\mathrm{x}}} \\ {^{\mathbf{B_M}}{f}_{\mathrm{y}}} \\ {^{\mathbf{B_M}}{f}_{\mathrm{z}}} \\ {^{\mathbf{B_M}}{m}_{\mathrm{x}}} \\ {^{\mathbf{B_M}}{m}_{\mathrm{y}}} \\ {^{\mathbf{B_M}}{m}_{\mathrm{z}}}
	\end{pmatrix},
\end{equation}
and thus the used notation.

Further, the  position vector of  frame $\left\lbrace \mathbf{B_M} \right\rbrace$ with respect to  frame $\left\lbrace \mathbf{B_C} \right\rbrace$, expressed in  $\left\lbrace \mathbf{B_C} \right\rbrace$ frame is:
\begin{equation}
	{^{\mathbf{B_C}}_{\mathbf{B_C}}{\boldsymbol{r}_{\mathbf{B_M}}}} = \begin{pmatrix}
		x_{\mathrm{CA}} \\ y_{\mathrm{CA}} \\ z_{\mathrm{CA}}
	\end{pmatrix},
\label{eqn: arm location}
\end{equation}
and the components of this vector account explicitly for the placement of the manipulator arm base and how this placement affects the supporting forces.

Chassis tensor of inertia can be assumed to have a general form with no zero elements, where the chassis moments of inertia about the $\left\lbrace \mathbf{B_C} \right\rbrace$ frame axes are denoted by ${^{\mathbf{B_C}}I_{\mathrm{ij}}} = {^{\mathbf{B_C}}I_{\mathrm{ji}}}, i,j = \mathrm{x, y, z}$, and participate in \eqref{eq: b1}--\eqref{eq: b3}.

In addition to the moments of inertia, the wheeled-platform mass has to be considered, and it is denoted by $m_{\mathrm{BC}}$. The chassis COM height above the ground, a known contributing term, is labelled as $z_{\rm C}$ and is also sketched in Fig. \ref{fig: vehicle and wheels frames}.

The lengths $l_1$ and $l_2$ have been already introduced in Fig. \ref{fig: platform kinematics} and addressed in Section \ref{sec: kinematic chain}. They represent distances from the chassis COM to the rear and front axle, respectively.

Finally, terms $r_{ij}$,where  $i,j=1,2,3$ are $ij$-th coefficients of the transformation matrix (\ref{eqn: GRBC}), and it is these terms that do account for the slope angles and thus for the chassis orientation with respect to the Earth-tangent plane. 

Solving the system of equations \eqref{eq: normal forces final eq}, compact expressions for each tyre's supporting force are obtained in the form of following neat sums with eleven terms in each: 
\begin{equation}
{^{\mathbf{T_{FL}}}f_z} = \displaystyle\sum_{i=1}^
{11} \zeta_{\mathrm{FL},i},
\label{eqn: TFLfz}
\end{equation}
\begin{equation}
	{^{\mathbf{T_{FR}}}f_z} = \displaystyle\sum_{i=1}^
	{11} \zeta_{\mathrm{FR},i},
\label{eqn: TFRfz}	
\end{equation}
\begin{equation}
	{^{\mathbf{T_{RR}}}f_z} = \displaystyle\sum_{i=1}^
	{11} \zeta_{\mathrm{RR},i},
\label{eqn: TRRfz}
\end{equation}
\begin{equation}
	{^{\mathbf{T_{RL}}}f_z} = \displaystyle\sum_{i=1}^
	{11} \zeta_{\mathrm{RL},i}.
\label{eqn: TRLfz}
\end{equation}

\newpage

Analytic expressions for all the contributing terms in \eqref{eqn: TFLfz}--\eqref{eqn: TRLfz} are given by \eqref{eqn: FL1}--\eqref{eqn: RL11}. Each term can be considered \textit{static} if its contribution exists irrespective of  the chassis or manipulator arm motion. Magnitudes of static terms are affected by the system masses, slope angles, manipulator arm posture, and the position of the connection point between the manipulator and the chassis. How the chassis and wheel weight forces are contributing to the supporting forces is quantified with \eqref{eqn: FL1} -- \eqref{eqn: RL1}. In this case, slope angles are  important contributors, together with the COM height.

\begin{strip}
\begin{equation}
	\zeta_{\mathrm{FL},1} = g  \, \left( \dfrac{m_{\mathrm{BC}}}{2} \dfrac{ l_1}{  l_1 + l_2 } + m_{\mathrm{w}} \right) \, r_{33} - \dfrac{g \, \left( m_{\rm w} \, R_{\rm w}  + 0.25 \, m_{\mathrm{BC}} \, z_{\rm C}\right)}{w} \, r_{32} - \dfrac{2 \, g \, \left( m_{\rm w} \, R_{\rm w} + 0.25 \, m_{\mathrm{BC}} \, z_{\rm C}  \right)}{l_1 + l_2} \, r_{31}
	\label{eqn: FL1}
\end{equation}
\begin{equation}
	\zeta_{\mathrm{FR},1} = g  \, \left( \dfrac{m_{\mathrm{BC}}}{2} \dfrac{ l_1}{  l_1 + l_2 } + m_{\mathrm{w}} \right) \, r_{33} + \dfrac{g \, \left( m_{\rm w} \, R_{\rm w}  + 0.25 \, m_{\mathrm{BC}} \, z_{\rm C}\right)}{w} \, r_{32} - \dfrac{2 \, g \, \left( m_{\rm w} \, R_{\rm w} + 0.25 \, m_{\mathrm{BC}} \, z_{\rm C}  \right)}{l_1 + l_2} \, r_{31}
\end{equation}
\begin{equation}
	\zeta_{\mathrm{RR},1} = g  \, \left( \dfrac{m_{\mathrm{BC}}}{2} \,\dfrac{l_2}{l_1 + l_2} + m_{\mathrm{w}} \right) \, r_{33} + \dfrac{g \, \left( m_{\rm w} \, R_{\rm w}  + 0.25 \, m_{\mathrm{BC}} \, z_{\rm C}\right)}{w} \, r_{32} + \dfrac{2 \, g \, \left( m_{\rm w} \, R_{\rm w} + 0.25 \, m_{\mathrm{BC}} \, z_{\rm C}  \right)}{l_1 + l_2} \, r_{31}
\end{equation}
\begin{equation}
	\zeta_{\mathrm{RL},1} = g  \, \left( \dfrac{m_{\mathrm{BC}}}{2} \, \dfrac{l_2}{l_1 + l_2 } + m_{\mathrm{w}} \right) \, r_{33} - \dfrac{g \, \left( m_{\rm w} \, R_{\rm w}  + 0.25 \, m_{\mathrm{BC}} \, z_{\rm C}\right)}{w} \, r_{32} + \dfrac{2 \, g \, \left( m_{\rm w} \, R_{\rm w} + 0.25 \, m_{\mathrm{BC}} \, z_{\rm C}  \right)}{l_1 + l_2} \, r_{31}
	\label{eqn: RL1}
\end{equation}
\end{strip}
It is \textit{stato-dynamic} terms \eqref{eqn: FL2} -- \eqref{eqn: RL6} that describe how the manipulator arm affects the normal loads. Location of the arm connection point with the chassis \eqref{eqn: arm location} plays an integral role in these terms. They can generally have non-zero static values, which change when motion exists. 
\begin{equation}	
		\zeta_{\mathrm{FL},2} = \dfrac{1}{2} \dfrac{ z_{\mathrm{CA}}  + z_{\mathrm{C}}}{l_1 + l_2} \, {^{\mathbf{B_M}}{f}_{\mathrm{x}}}
		\label{eqn: FL2} 	
\end{equation}
\begin{equation}
	\zeta_{\mathrm{FR},2} = \dfrac{1}{2}  \, \dfrac{ z_{\mathrm{CA}}  + z_{\mathrm{C}}  }{ l_1 + l_2 } \, {^{\mathbf{B_M}}{f}_{\mathrm{x}}}
\end{equation}
\begin{equation}
		\zeta_{\mathrm{RR},2} = - \dfrac{1}{2}   \, \dfrac{ z_{\mathrm{CA}}  + z_{\mathrm{C}} }{ l_1 + l_2}  \, {^{\mathbf{B_M}}{f}_{\mathrm{x}}}
\end{equation}
\begin{equation}
		\zeta_{\mathrm{RL},2} = - \dfrac{1}{2}  \, \dfrac{ z_{\mathrm{CA}}  + z_{\mathrm{C}} }{ l_1 + l_2 }  \, {^{\mathbf{B_M}}{f}_{\mathrm{x}}}
\end{equation}
\begin{equation}
		\zeta_{\mathrm{FL},3}  = \dfrac{1}{4} \, \dfrac{ z_{\mathrm{CA}}  + z_{\mathrm{C}} }{w} \, {^{\mathbf{B_M}}{f}_{\mathrm{y}}}  
\end{equation}
\begin{equation}
	\zeta_{\mathrm{FR},3} =- \dfrac{1}{4} \, \dfrac{ z_{\mathrm{CA}}  + z_{\mathrm{C}}  }{w} \, {^{\mathbf{B_M}}{f}_{\mathrm{y}}}  
\end{equation}
\begin{equation}
		\zeta_{\mathrm{RR},3} =- \dfrac{1}{4} \, \dfrac{ z_{\mathrm{CA}}  + z_{\mathrm{C}} }{ w} \, {^{\mathbf{B_M}}{f}_{\mathrm{y}}}  
\end{equation}
\begin{equation}
		\zeta_{\mathrm{RL},3} = \dfrac{1}{4} \, \dfrac{ z_{\mathrm{CA}}  + z_{\mathrm{C}}}{ w} \, {^{\mathbf{B_M}}{f}_{\mathrm{y}}}  
\end{equation}
\begin{equation}
		\zeta_{\mathrm{FL},4} =  - \dfrac{1}{2} \, \left( \, \dfrac{l_1 + x_{\mathrm{CA}}}{ l_1 + l_2 }  + \dfrac{1}{2} \,\dfrac{y_{\mathrm{CA}}}{w} \right) \, {^{\mathbf{B_M}}{f}_{\mathrm{z}}}
\end{equation}
\begin{equation}
	\zeta_{\mathrm{FR},4} = -\dfrac{1}{2} \, \left(    \dfrac{l_1 + x_{\mathrm{CA}}}{ l_1 + l_2 }  - \dfrac{1}{2} \,\dfrac{y_{\mathrm{CA}}}{w} \right) \, {^{\mathbf{B_M}}{f}_{\mathrm{z}}}
\end{equation}
\begin{equation}
		\zeta_{\mathrm{RR},4} =  - \dfrac{1}{2} \, \left( \, \dfrac{l_2 - x_{\mathrm{CA}}}{ l_1 + l_2 }  - \dfrac{1}{2} \,\dfrac{y_{\mathrm{CA}}}{w} \right) \, {^{\mathbf{B_M}}{f}_{\mathrm{z}}}
\end{equation}
\begin{equation}
		\zeta_{\mathrm{RL},4} =  - \dfrac{1}{2} \, \left( \, \dfrac{l_2 - x_{\mathrm{CA}}}{ l_1 + l_2 }  + \dfrac{1}{2} \,\dfrac{y_{\mathrm{CA}}}{w} \right) \, {^{\mathbf{B_M}}{f}_{\mathrm{z}}}
\end{equation}
\begin{equation}
		\zeta_{\mathrm{FL},5}  =- \dfrac{{^{\mathbf{B_M}}{m}_{\mathrm{x}}} }{4 \, w} 
\end{equation}
\begin{equation}
	\zeta_{\mathrm{FR},5}  = \dfrac{{^{\mathbf{B_M}}{m}_{\mathrm{x}}} }{4 \, w}
\end{equation}
\begin{equation}
		\zeta_{\mathrm{RR},5}  =  \dfrac{{^{\mathbf{B_M}}{m}_{\mathrm{x}}} }{4 \, w} 
\end{equation}
\begin{equation}
		\zeta_{\mathrm{RL},5}  =- \dfrac{{^{\mathbf{B_M}}{m}_{\mathrm{x}}} }{4 \, w} 
\end{equation}
\begin{equation}
		\zeta_{\mathrm{FL},6} = \dfrac{1}{2} \, \dfrac{{^{\mathbf{B_M}}{m}_{\mathrm{y}}} }{l_1 + l_2} 
\end{equation}
\begin{equation}
	\zeta_{\mathrm{FR},6} = \dfrac{1}{2} \, \dfrac{{^{\mathbf{B_M}}{m}_{\mathrm{y}}} }{l_1 + l_2 }
\end{equation}
\begin{equation}
		\zeta_{\mathrm{RR},6} = -\dfrac{1}{2} \, \dfrac{{^{\mathbf{B_M}}{m}_{\mathrm{y}}} }{l_1 + l_2} 
\end{equation}
\begin{equation}
		\zeta_{\mathrm{RL},6} = -\dfrac{1}{2} \, \dfrac{{^{\mathbf{B_M}}{m}_{\mathrm{y}}} }{l_1 + l_2 } 
		\label{eqn: RL6}
\end{equation}

Terms \eqref{eqn: FL7} -- \eqref{eqn: RL11} exist only when the wheeled platform is moving, and thus can be addressed as \textit{dynamic} terms. These provide insight into how the chassis linear/angular velocities and accelerations affect the normal wheel loads.
\begin{equation}
	\zeta_{\mathrm{FL},7} =  -\dfrac{m_{\mathrm{BC}}}{2} \dfrac{ z_{\mathrm{C}} }{ l_1 + l_2} \, {^{\mathbf{B_C}}\dot{v}_{\mathrm{x}}}
	\label{eqn: FL7}
\end{equation}
\begin{equation}
	\zeta_{\mathrm{FR},7} =  -\dfrac{m_{\mathrm{BC}}}{2} \dfrac{ z_{\mathrm{C}} }{ l_1 + l_2} \, {^{\mathbf{B_C}}\dot{v}_{\mathrm{x}}}
\end{equation}
\begin{equation}
	\zeta_{\mathrm{RR},7} =   \dfrac{m_{\mathrm{BC}}}{2} \dfrac{ z_{\mathrm{C}} }{ l_1 + l_2} \, {^{\mathbf{B_C}}\dot{v}_{\mathrm{x}}}
\end{equation}
\begin{equation}
	\zeta_{\mathrm{RL},7} =   \dfrac{m_{\mathrm{BC}}}{2} \dfrac{ z_{\mathrm{C}} }{ l_1 + l_2} \, {^{\mathbf{B_C}}\dot{v}_{\mathrm{x}}}
\end{equation}
\begin{equation}
		\zeta_{\mathrm{FL},8} =  - \dfrac{m_{\mathrm{BC}}}{4} \dfrac{z_{\mathrm{C}} }{w} \, {^{\mathbf{B_C}}\dot{v}_{\mathrm{y}}} 
\end{equation}
\begin{equation}
		\zeta_{\mathrm{FR},8} =   \dfrac{m_{\mathrm{BC}}}{4} \dfrac{z_{\mathrm{C}} }{w} \, {^{\mathbf{B_C}}\dot{v}_{\mathrm{y}}} 
\end{equation}
\begin{equation}
		\zeta_{\mathrm{RR},8} =   \dfrac{m_{\mathrm{BC}}}{4} \dfrac{z_{\mathrm{C}} }{w} \, {^{\mathbf{B_C}}\dot{v}_{\mathrm{y}}} 
\end{equation}
\begin{equation}
		\zeta_{\mathrm{RL},8} = -  \dfrac{m_{\mathrm{BC}}}{4} \dfrac{z_{\mathrm{C}} }{w} \, {^{\mathbf{B_C}}\dot{v}_{\mathrm{y}}} 
\end{equation}
\begin{equation}
		\zeta_{\mathrm{FL},9} = \dfrac{m_{\mathrm{BC}}}{2} \, z_{\mathrm{C}} \,  \left( -\dfrac{{^{\mathbf{B_C}}{v}_{\mathrm{x}}}}{2 \, w}   + \dfrac{{^{\mathbf{B_C}}{v}_{\mathrm{y}}}}{ l_1 + l_2 } \right) \,  \dot{\Psi}
		\label{eq: FL9}
\end{equation}
\begin{equation}
		\zeta_{\mathrm{FR},9} = \dfrac{m_{\mathrm{BC}}}{2} \, z_{\mathrm{C}} \,  \left( \dfrac{{^{\mathbf{B_C}}{v}_{\mathrm{x}}}}{2 \, w}   + \dfrac{{^{\mathbf{B_C}}{v}_{\mathrm{y}}}}{ l_1 + l_2 } \right) \,  \dot{\Psi}
\end{equation}
\begin{equation}
		\zeta_{\mathrm{RR},9} = - \dfrac{m_{\mathrm{BC}}}{2} \, z_{\mathrm{C}} \,  \left( -\dfrac{{^{\mathbf{B_C}}{v}_{\mathrm{x}}}}{2 \, w}   + \dfrac{{^{\mathbf{B_C}}{v}_{\mathrm{y}}}}{ l_1 + l_2 } \right) \,  \dot{\Psi}
\end{equation}
\begin{equation}
		\zeta_{\mathrm{RL},9} = - \dfrac{m_{\mathrm{BC}}}{2} \, z_{\mathrm{C}} \,  \left( \dfrac{{^{\mathbf{B_C}}{v}_{\mathrm{x}}}}{2 \, w}   + \dfrac{{^{\mathbf{B_C}}{v}_{\mathrm{y}}}}{ l_1 + l_2 } \right) \,  \dot{\Psi}
		\label{eq: RL9}
\end{equation}
\begin{equation}
		\zeta_{\mathrm{FL},10} = -\dfrac{1}{2}  \,  \left( \dfrac{{^{\mathbf{B_C}}{I}_{\mathrm{xz}}}}{ l_1 + l_2 }   + \dfrac{{^{\mathbf{B_C}}{I}_{\mathrm{yz}}}}{2     \,  w} \right) \,  \dot{\Psi}^2
\end{equation}
\begin{equation}
		\zeta_{\mathrm{FR},10} = -\dfrac{1}{2}  \,  \left( \dfrac{{^{\mathbf{B_C}}{I}_{\mathrm{xz}}}}{ l_1 + l_2 }   - \dfrac{{^{\mathbf{B_C}}{I}_{\mathrm{yz}}}}{2     \,  w} \right) \,  \dot{\Psi}^2
\end{equation}
\begin{equation}
		\zeta_{\mathrm{RR},10} = \dfrac{1}{2}  \,  \left( \dfrac{{^{\mathbf{B_C}}{I}_{\mathrm{xz}}}}{ l_1 + l_2 }   + \dfrac{{^{\mathbf{B_C}}{I}_{\mathrm{yz}}}}{2     \,  w} \right) \,  \dot{\Psi}^2
\end{equation}
\begin{equation}
		\zeta_{\mathrm{RL},10} = \dfrac{1}{2}  \,  \left( \dfrac{{^{\mathbf{B_C}}{I}_{\mathrm{xz}}}}{ l_1 + l_2 }   - \dfrac{{^{\mathbf{B_C}}{I}_{\mathrm{yz}}}}{2     \,  w} \right) \,  \dot{\Psi}^2
\end{equation}
\begin{equation}
		\zeta_{\mathrm{FL},11} = - \dfrac{1}{2} \, \left( \dfrac{{^{\mathbf{B_C}}{I}_{\mathrm{yz}}}}{ l_1 + l_2 }   - \dfrac{{^{\mathbf{B_C}}{I}_{\mathrm{xz}}}}{2     \,  w} \right) \,  \ddot{\Psi}
\end{equation}
\begin{equation}
		\zeta_{\mathrm{FR},11} = - \dfrac{1}{2} \, \left( \dfrac{{^{\mathbf{B_C}}{I}_{\mathrm{yz}}}}{ l_1 + l_2 }   + \dfrac{{^{\mathbf{B_C}}{I}_{\mathrm{xz}}}}{2     \,  w} \right) \,  \ddot{\Psi}
\end{equation}
\begin{equation}
		\zeta_{\mathrm{RR},11} =  \dfrac{1}{2} \, \left( \dfrac{{^{\mathbf{B_C}}{I}_{\mathrm{yz}}}}{ l_1 + l_2 }   - \dfrac{{^{\mathbf{B_C}}{I}_{\mathrm{xz}}}}{2     \,  w} \right) \,  \ddot{\Psi}
\end{equation}
\begin{equation}
		\zeta_{\mathrm{RL},11} =  \dfrac{1}{2} \, \left( \dfrac{{^{\mathbf{B_C}}{I}_{\mathrm{yz}}}}{ l_1 + l_2 }   + \dfrac{{^{\mathbf{B_C}}{I}_{\mathrm{xz}}}}{2     \,  w} \right) \,  \ddot{\Psi}
		\label{eqn: RL11}
\end{equation}

\section{Simulation results}

\label{sec: simulation}

The presented derivation procedure and consequently the solutions for normal wheel loads have been based on the basic principles of rigid body dynamics with several reasonable assumptions introduced in the derivation process. As an external independent means for a self check-up, the Simscape Multibody\texttrademark \, simulation will be used here to advocate the proposed modelling scheme's justifiability and provide insight into how the introduced assumptions affect the final solution when they are not entirely valid. A 4-AWD mobile manipulator in the simulation environment can be freely chosen since equations were not tailored to any specific chassis shape or manipulator arm and their properties.

It must be noted that there exists a temporary inability to provide referent normal force values on an uneven, complex geometry terrain in the software used. State-of-the-art methods used to simulate a motion over an uneven terrain use clouds of points. Each contact force is based on the penetration and velocity of the individual point of the cloud. It must be noted that the Spatial Contact Force block does not support sensing when connected to a Point Cloud block, \cite{b35}.
This fact has forced simulations here to be performed on a flat surface for the sake of providing an unbiased self-check comparison. Even these will suffice to show the benefits of monitoring wheel loads instead of tipping-over moments about pre-defined axes. In the case of uneven terrain, the argumentation brought out during the derivation process provides a firm basis in the temporary absence of unbiased self-check means.

When the motion of a mobile manipulator is simulated over a flat surface, a simple spring-damper model for the soil can be used without caring about the terrain geometry, as suggested and implemented in \cite{b36}. This way, reference

\Figure[t!](topskip=0pt, botskip=0pt, midskip=0pt)[width=.83\linewidth]{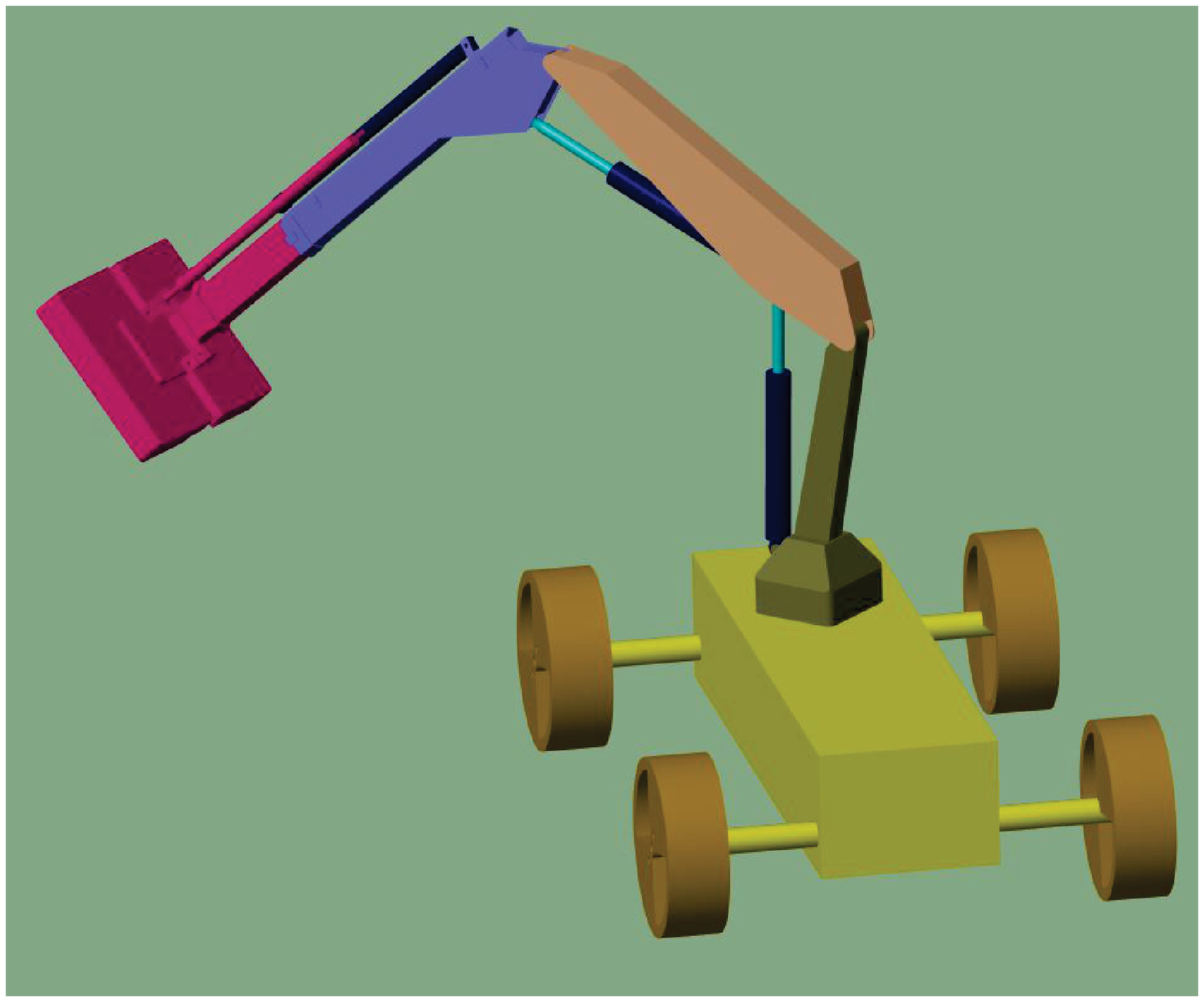}
{The simplified mobile manipulator in Simscape Multibody\texttrademark \, simulation environment used to justify the proposed modelling concept. \label{fig7}}
  
\Figure[t!](topskip=0pt, botskip=0pt, midskip=0pt)[width=.99\linewidth]{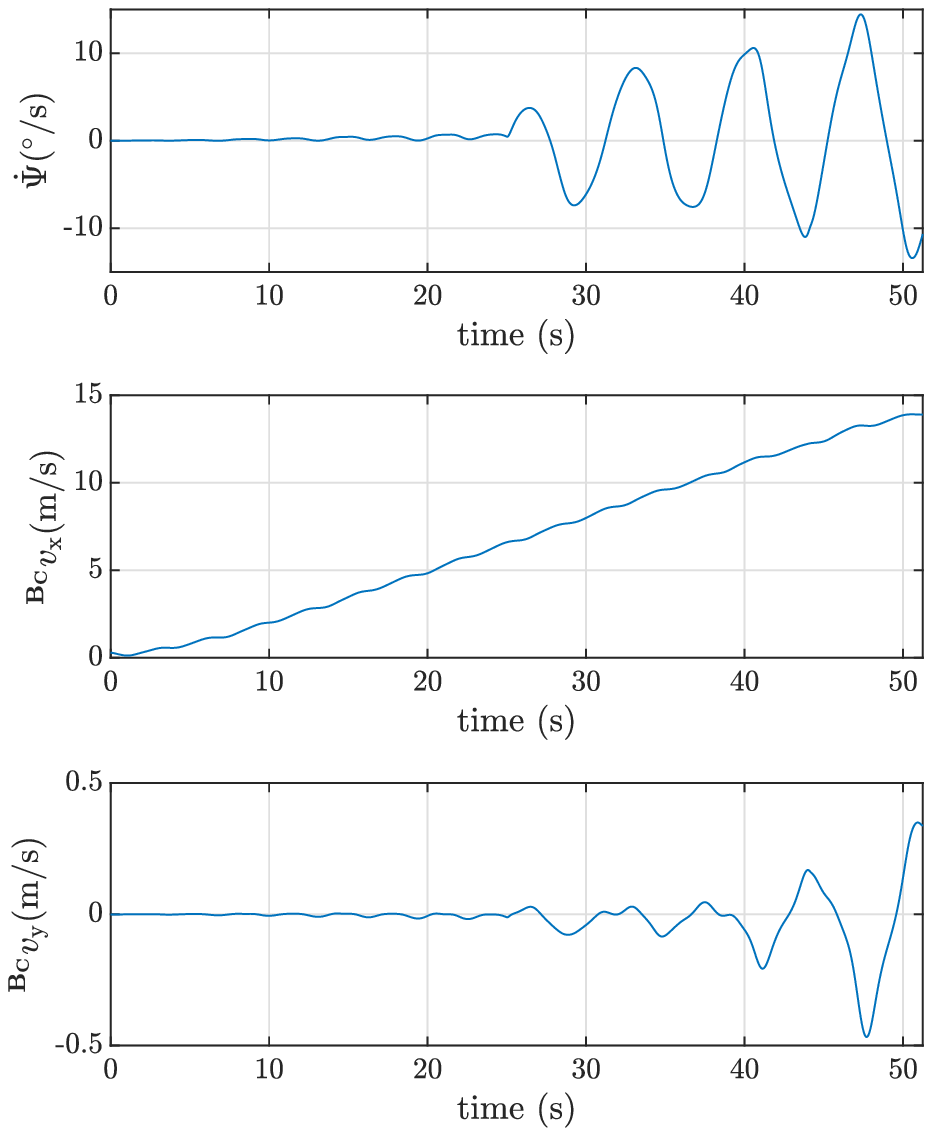}
{Linear/angular chassis velocities in the simulation. \label{fig8}}

\begin{figure*}[htp]
	\centering
	\includegraphics[width=\textwidth]{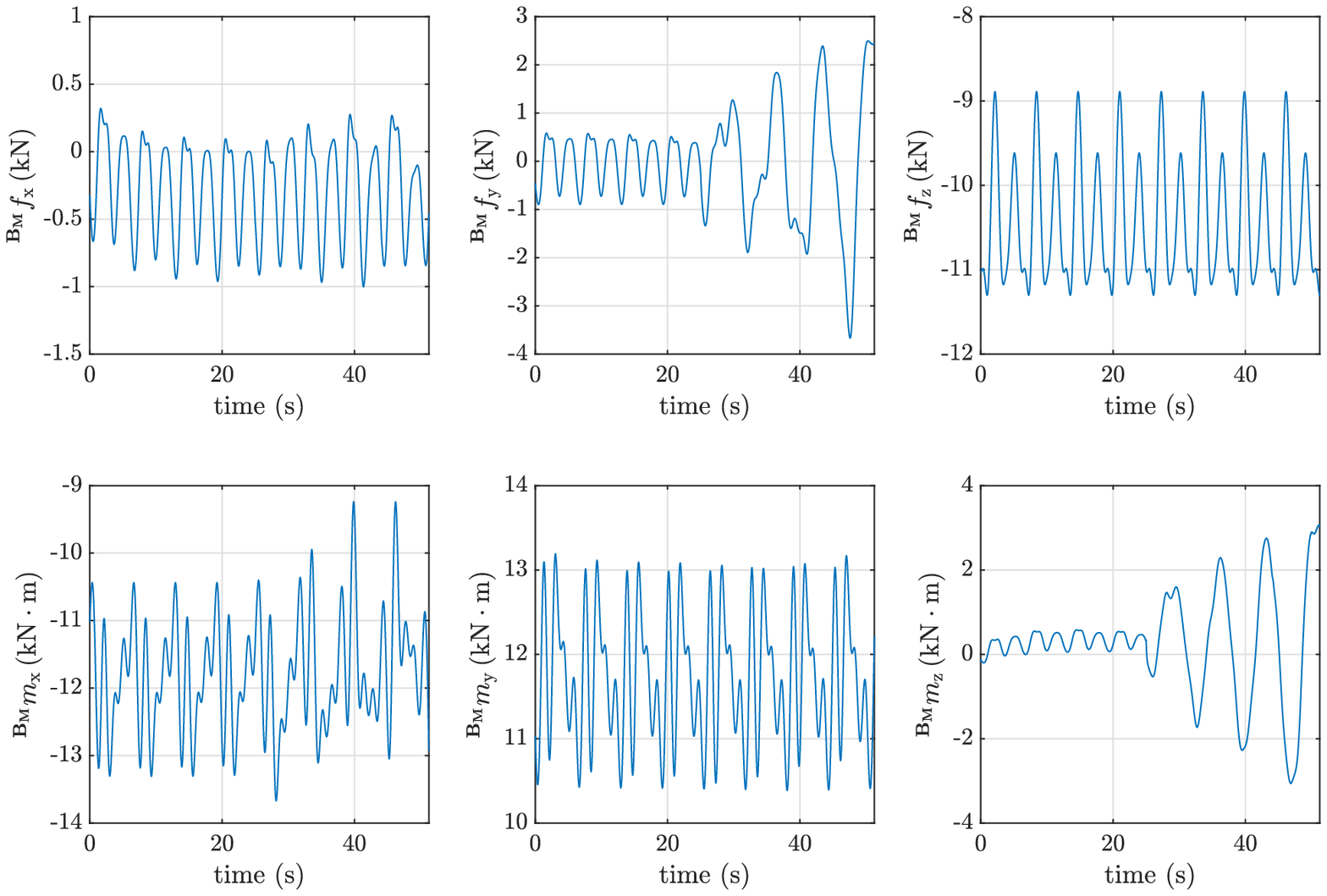}
	\caption{Forces and moments at the chassis/manipulator arm connection point, expressed as would have been measured in $\left\lbrace \mathbf{B_M} \right\rbrace$ frame.} \label{fig: fig9}
\end{figure*}

\newpage
 
\noindent  values for the result comparisons and proof-of-concept purposes can be obtained relatively easy. 

Flat-surface simulations are, in any case, significant when testing both static and dynamic terms at the same time. It is also expected that in the case of large angular velocities, more significant discrepancies between the referent results and the analytic ones will exist due to introduced assumptions that neglect the wheel dynamics. These cases can be realistically simulated only on flat terrain. On uneven terrain, static terms will usually prevail because the manipulator will not have high linear/angular velocities in those cases. 

The heavy-duty mobile manipulator shown in Fig. \ref{fig7} will be used in the simulation. The manipulator arm will intentionally perform motions with significant angular accelerations to create ${^{\mathbf{B_M}}{\boldsymbol{F}}}$ components of a considerable magnitude. The simulation may exaggerate a situation that is likely to happen in practice but simultaneously tests the static and dynamic factors for a wider spectre of affecting values. All the mass-related and other relevant physical properties of the manipulator arm can be found in \cite{b37}, together with the detailed description of the forces calculation procedures. Other relevant simulation parameters are listed in Table \ref{table}.
\begin{table}[h]
	\caption{Wheeled platform parameters}
	\label{table}
	\setlength{\tabcolsep}{3pt}
	\begin{tabular}{|p{25pt}|p{130pt}|p{30pt}|p{30pt}|}
		\hline
		Label& 
		Quantity& 
		Value & Unit \\
		\hline
		$m_{\mathrm{BC}}$& 
		chassis mass& 
		2200 & kg\\
		$l_1$& 
		longitudinal distance between the chassis COM and rear axle& 
		 1.15 & m\\
		$l_2$& 
		longitudinal distance between the chassis COM and front axle& 
		 1.15 & m\\
		$w$& 
		lateral distance between the chassis COM and all the wheels& 
		 0.875&m\\
		$m_{\rm w}$& 
		mass of the each wheel& 60
		& kg\\
		$R_{\rm w}$& 
		radius of the each wheel& 
		0.35&m\\
		$z_{\rm C}$& 
		chassis COM height above the ground& 
		0.45& m\\
		$x_{\mathrm{CA}}$& 
		$x$-axis distance from \eqref{eqn: arm location}& 0.5
		 &m\\
		$y_{\mathrm{CA}}$& 
		$y$-axis distance from \eqref{eqn: arm location}& 0
		& m\\
		$z_{\mathrm{CA}}$& 
		$z$-axis distance from \eqref{eqn: arm location}& 0.25
		 &m\\
		$g$& 
		gravity acceleration& 9.8066
	    & m/$\rm s^2$\\
		\hline
	\end{tabular}
	\label{tab1}
\end{table}

The manipulator arm has also been intentionally oriented towards the front left wheel such that the rear right wheel is in the greatest danger of losing ground contact. This occurrence is one of those that are not likely to be detected using the existing indicators, and thus it is of primary interest here with a goal in mind to reveal how the proposed approach prevails the ITOM and its predecessors. 

Linear/angular chassis velocities in the simulated case are shown in Fig. \ref{fig8}. Forces/moments generated at the manipulator arm/chassis connection ${^{\mathbf{B_M}}{\boldsymbol{F}}}$ are shown in Fig. \ref{fig: fig9}. Initially, these forces and moments are mainly caused by the manipulator arm motion and have been afterwards magnified by abrupt steering actions that are purposedly introduced to test the contribution of angular velocity. It must be emphasised that all the motions were entirely mutually uncorrelated.

\begin{figure*}[htp]
	\centering
	\includegraphics[width=.95\textwidth]{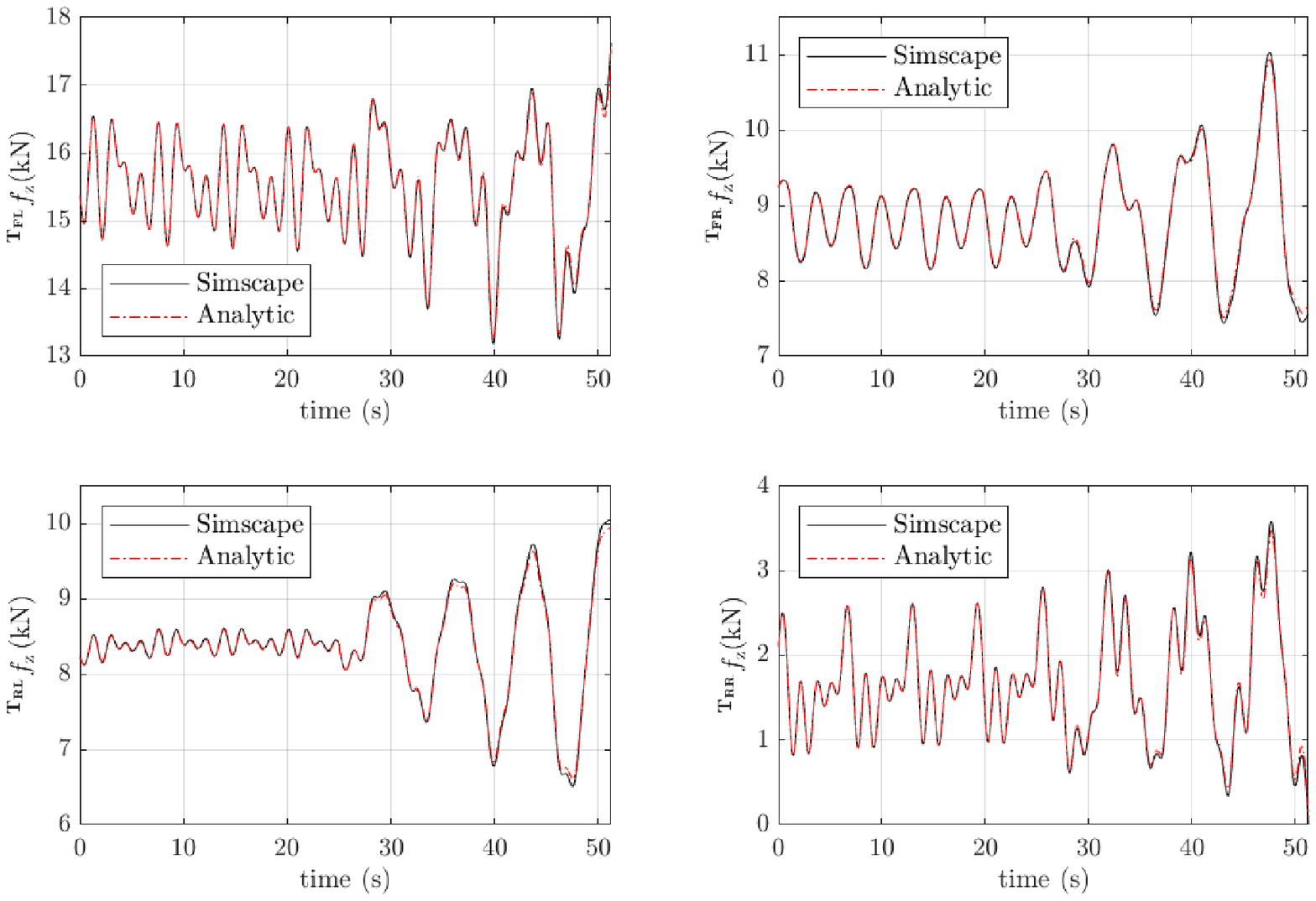}
	\caption{Wheel supporting forces on all the wheels. Analytical results are compared with referent values from the Simscape Multibody\texttrademark. The tip-over danger can be detected most properly by monitoring the supporting forces separately.} \label{fig: normal reaction in sim}
\end{figure*}
\begin{figure*}[htp]
	\centering
	\includegraphics[width=.95\textwidth]{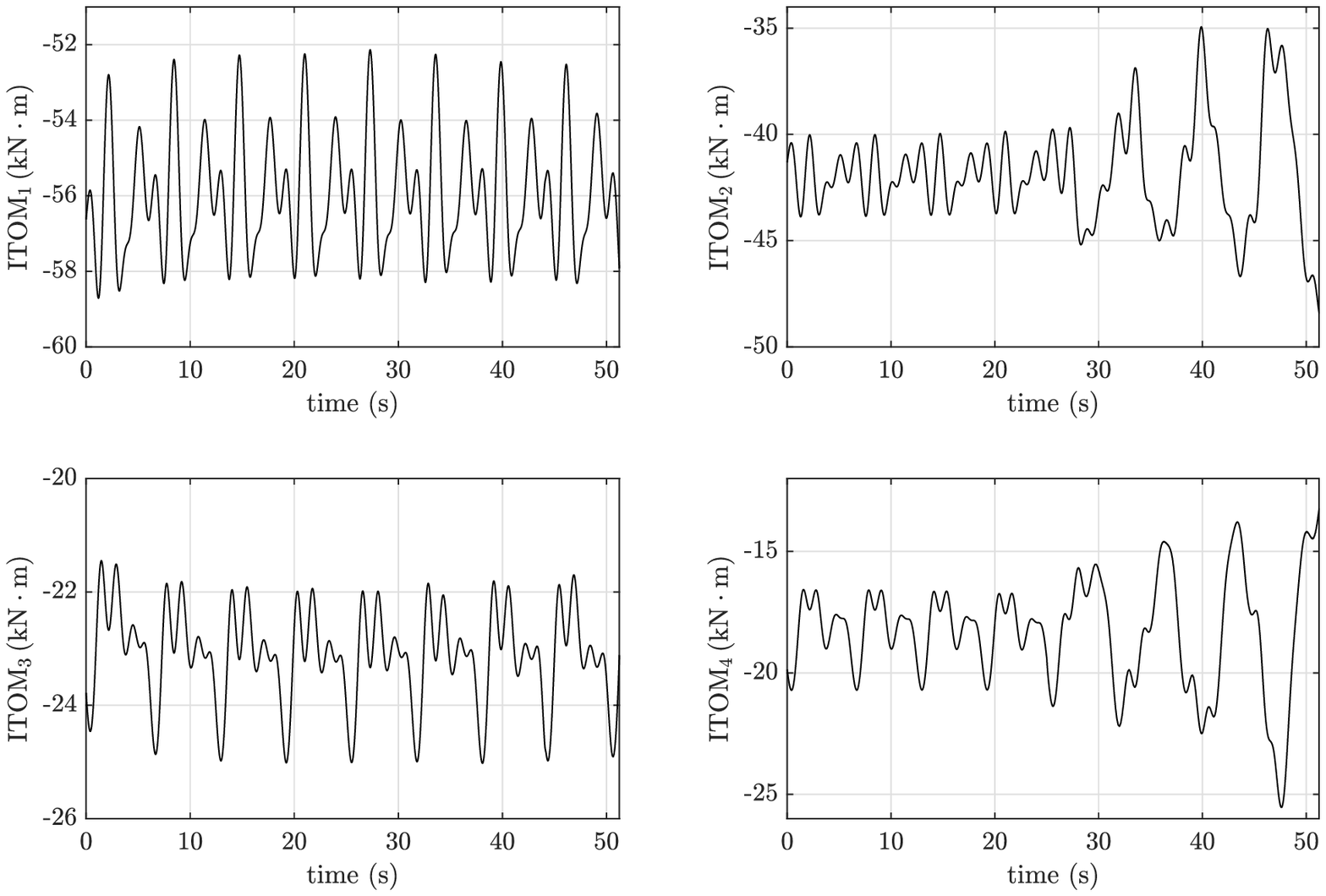}
	\caption{Values of the ITOM tip-over stability indicator. Negative ITOM values designate the tip-over stability. By considering only tipping-over about the axes connecting wheels, it cannot detect the case when only one wheel loses contact with the ground.} \label{fig11}
\end{figure*}

\newpage

Values of the supporting forces calculated per the proposed approach, using \eqref{eqn: TFLfz} -- \eqref{eqn: TRLfz} are shown in Fig. \ref{fig: normal reaction in sim}, and these results are significant from two viewpoints. 

First of all, they show an excellent agreement between the simulation and analytic results. This fact approves using the proposed extendable model with valid underlying assumptions.  The proposed equations qualify as a great starting point in calculating the supporting forces considering the obtained matching with the Simscape Multibody\texttrademark \, results, and that simulation has been carried out neutrally. Although the discrepancies between the analytic and simulation results depend on more than introduced assumptions, as on the solver choice, integration time, soil model, and similar, for the investigated manoeuvre, the maximum absolute error is negligible at first, having the maximum value about 130$\rm N$ as the motion becomes sharp, which is arguably more than acceptable considering the magnitudes of forces. End-user could investigate dissimilarities from the referent results and decide if a more complex model is required. In the absence of referent simulation or experimental results, the given methodology can be adopted in all the practical situations involving the considered class of mobile manipulators. 

From the second standpoint, improvement with respect to the ITOM is noticeable. Namely, by observing the values for ${^{\mathbf{T_{RR}}}f_z}$, it can be seen that the rear right wheel loses contact with the surface at the point in time when the simulation ends. The convention in the ITOM tip-over stability indicator is such that negative values imply the tip-over stability. The transition from negative to positive ITOM values can occur only when the two wheels lose contact with the ground. As shown in Fig. \ref{fig11}, although a positive gradient in some may exist, even the highest ITOM indicator value is far below zero, indicating strong tip-over stability when the rear right wheel loses  contact with the ground. 

Monitoring the supporting forces per the suggested equations is intuitive and proved to be a better choice than monitoring the tipping-over moments about particular axes, no matter how detailed the underlying dynamics model is. Tip-over Force (TOF) stability measure emerges smoothly from the ongoing discussion as a simple alternative to ITOM. In contrast to the established procedures of comparing tipping-over moments, a comparison of the tipping-over forces with the prescribed stability margins is proposed. Tipping-over force is simply a wheel supporting force:
\begin{equation}
	\mathrm{TOF}_i = {^{\mathbf{T_{i}}}f_z},
\end{equation}
where $i = \mathrm{FL, FR, RR, RL}$. For a reasonably chosen optimal margins $\mathrm{TOF}_{i, \mathrm{opt}}$ one can now similarly write:
\begin{equation}
	\mathrm{TOF}_i = \mathrm{TOF}_{i, \mathrm{opt}} + \Delta \mathrm{TOF}_i.
\end{equation}

The tip-over avoidance function from \cite{b19} can be easily swapped with:
\begin{equation}
	\sigma = \dfrac{1}{2} \displaystyle\sum_i\left\lvert  \left\lvert \Delta \mathrm{TOF}_i \right\rvert \right\rvert_2^2,
\end{equation}
$i = \mathrm{FL, FR, RR, RL}$.  This introduces a great benefit of having analytically partial derivatives $\dfrac{\partial \sigma}{\partial \xi}$ used in the the tip-over avoidance scheme already proven to work, with $\xi = \begin{pmatrix}
	{^{\mathbf{B_C}}{v}_{\mathrm{x}}} &  {^{\mathbf{B_C}}{v}_{\mathrm{y}}} & \dot{\Psi}
\end{pmatrix}^T$.  In the light of performed analysis, it is also more relevant to choose a $\mathrm{TOF}_{i, \mathrm{max}}$ margin.

\section{Discussion}

\label{sec: discussion}

Analytic results for normal wheel loads and their derivation have so far stayed off the radar in the case of a wheeled mobile manipulator for an unknown reason. Although otherwise may be argued, obtaining these expressions does not pose a significant challenge if the line of thought commonly employed in vehicle dynamics is followed. After the careful subsystem-by-subsystem dynamics formulation, using the compact 6D vector form, which explicitly includes the ground reaction forces,  an appropriate system of equations with tyre loads as unknowns can be formed. The EOMs in which normal loads participate are initially hard to handle and provide highly impractical solutions. An elegant, easy-to-solve system is obtained by carefully combining these with the remaining EOMs. With reasonable assumptions, the solution for each normal force is a neat sum of 11 terms that accurately captures the normal load changes. This sum can be potentially further simplified or made even more complex from case to case if required. An independent comparison of the obtained analytical results with results obtained using Simscape Multibody\texttrademark \, for one random wheeled platform with a heavy-duty serial-parallel manipulator on top, advocates the proposed equations' adequacy.

In contrast to the existing tipping-over stability criteria, from now on, normal loads can be relatively easily imported to tip-over stability analysis. This new opportunity of inclusion offers serious effectivity. The assumptions on the tipping-over axis can be removed once for all. A simple consequence is that all the possible underlying causes are comprehended when working in terms of forces. Also, it is intuitively clear that prescribing relative stability margins in terms of forces is more accessible and straightforward than overturning moments about different axes, which brings the derived results close to the broader audience of readers and end-users.

\section{Conclusion}

\label{sec: conclusions}

With the closed gap between the car dynamics and mobile manipulators in the analytical determination of the wheel supporting forces, new tipping-over criteria and tipping-over avoidance schemes can be formed. 

The purpose of the present study is to address the work previously done comprehensively and to provide modifiable analytic expressions for what seems to be a logical continuation of ongoing efforts in the research community.

Discussion of an uneven terrain effect in more detail calls for investigating tipping-over stability indicators, which also emphasise the terrain configuration since these may present an exciting and significant improvement to the existing ones.

\section*{Acknowledgment}

The authors express their gratitude to  Dr Janne Koivum$\ddot{\rm a}$ki and Lionel Hulttinen for providing positive criticism and constructive feedback on improving the paper contents and the material presentation.

\begin{IEEEbiography}[{\includegraphics[width=1in,height=1.25in,clip,keepaspectratio]{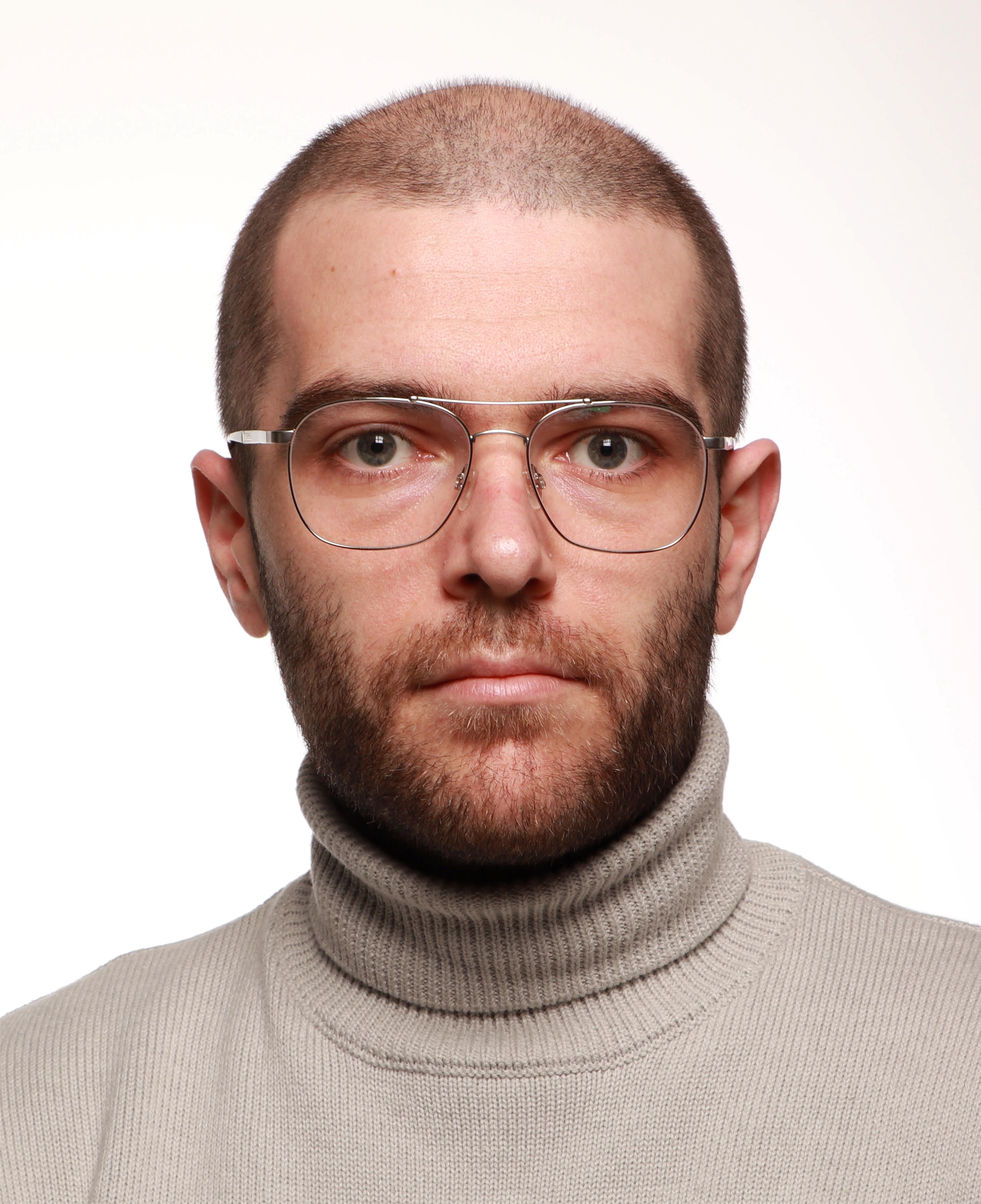}}]{Goran R. Petrovi{\'c}}  was born in Novi Pazar, SFR Yugoslavia in 1990. He received the B.S. and M.S. degrees in control engineering from the University of Belgrade, Faculty of Mechanical Engineering, Republic of Serbia, in 2014.
	
From March 2015 to March 2021, he was a Teaching Assistant with the Control Group at the University of Belgrade, Faculty of Mechanical Engineering. Since April 2021, he has been a Doctoral Researcher with the	Tampere University, Faculty of Engineering and Natural Sciences, Unit of Automation Technology and Mechanical Engineering, with prof. Jouni Mattila acting as a supervisor. His research interests include mobile manipulators, electro-hydraulic actuation and nonlinear model-based control algorithms. 
	
Mr. Petrovi{\'c} graduated as the best in his class and was a recipient of the Serbian Ministry of youth and sports scholarship "Dositeja", awarded to the 1000 best students in the Republic of Serbia.

\end{IEEEbiography}

\begin{IEEEbiography}[{\includegraphics[width=1in,height=1.25in,clip,keepaspectratio]{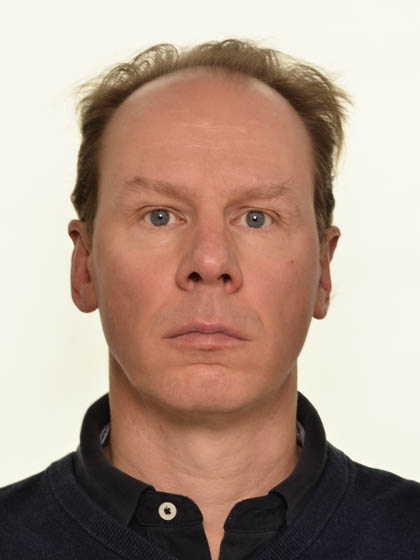}}]{Jouni Mattila} Dr. Tech. has received
	M.Sc. (Eng.) in 1995 and Dr. Tech. in 2000, both from the Tampere University of Technology (TUT),	Tampere, Finland. 
	
	He is currently a Professor of machine automation with the unit of Automation Technology and Mechanical Engineering, Tampere University. His research interests include machine automation, nonlinear model-based control of robotic manipulators, and energy-efficient control of heavy-duty mobile manipulators.
\end{IEEEbiography}

\EOD

\end{document}